\documentclass[lettersize,journal]{IEEEtran}
\usepackage{amsmath,amsfonts}
\usepackage{algorithmic}
\usepackage{algorithm}
\usepackage{array}
\usepackage{textcomp}
\usepackage{stfloats}
\usepackage{url}
\usepackage{subcaption}
\usepackage{verbatim}
\usepackage{hyperref}
\usepackage{graphicx}
\usepackage{cite}
\usepackage[table]{xcolor}

\begin{document}

\title{MMFusion: Combining Image Forensic Filters for Visual Manipulation Detection and Localization}

\author{
    \IEEEauthorblockN{Kostas Triaridis\IEEEauthorrefmark{1}\IEEEauthorrefmark{2}, Konstantinos Tsigos\IEEEauthorrefmark{1} and Vasileios Mezaris\IEEEauthorrefmark{1} }\\
    \IEEEauthorblockA{\IEEEauthorrefmark{1}Centre for Research and Technology Hellas (CERTH) / Information Technologies Institute (ITI) \\ Thermi 57001, Greece, \{triaridis, ktsigos, bmezaris\}@iti.gr}\\
    \IEEEauthorblockA{\IEEEauthorrefmark{2}Department of Computer Science, Stony Brook University, Stony Brook, NY 11794, USA \\ ktriaridis@cs.stonybrook.edu}\\
}

\maketitle

\begin{abstract}
Recent image manipulation localization and detection techniques typically leverage forensic artifacts and traces that are produced by a noise-sensitive filter, such as SRM or Bayar convolution. In this paper, we showcase that different filters commonly used in such approaches excel at unveiling different types of manipulations and provide complementary forensic traces. Thus, we explore ways of combining the outputs of such filters to leverage the complementary nature of the produced artifacts for performing image manipulation localization and detection (IMLD). We assess two distinct combination methods: one that produces independent features from each forensic filter and then fuses them (this is referred to as late fusion) and one that performs early mixing of different modal outputs and produces combined features (this is referred to as early fusion). We use the latter as a feature encoding mechanism, accompanied by a new decoding mechanism that encompasses feature re-weighting, for formulating the proposed MMFusion architecture. We demonstrate that MMFusion achieves competitive performance for both image manipulation localization and detection, outperforming state-of-the-art models across several image and video datasets. We also investigate further the contribution of each forensic filter within MMFusion for addressing different types of manipulations, building on recent AI explainability measures.
\end{abstract}


\maketitle

\section{Introduction}
\label{sec:introduction}
 Editing and manipulating digital media has gotten increasingly easier and more accessible in recent years. Recent advances in image editing software, as well as deep generative models such as Generative Adversarial Networks (GANs)\cite{lahiri2020prior, zhang2022gan} and diffusion models\cite{nichol2022glide, xie2023smartbrush}, facilitate producing manipulations that are often imperceptible to the human eye and are widely available, even to potentially malicious users. The widespread use of smartphones and social networks also enables the spread of such manipulated media at a rapid pace. As a result, such edited images can cause social problems when used as evidence to support disinformation campaigns and stories or mislead the public by obfuscating important content from news, resulting in diminished trust. Therefore, techniques for image manipulation detection and localization, as part of complete toolboxes for media verification such as \cite{teyssou2017invid}, are now needed more than ever. 

\begin{figure*}[t]
\centering
\includegraphics[width=0.7\textwidth]{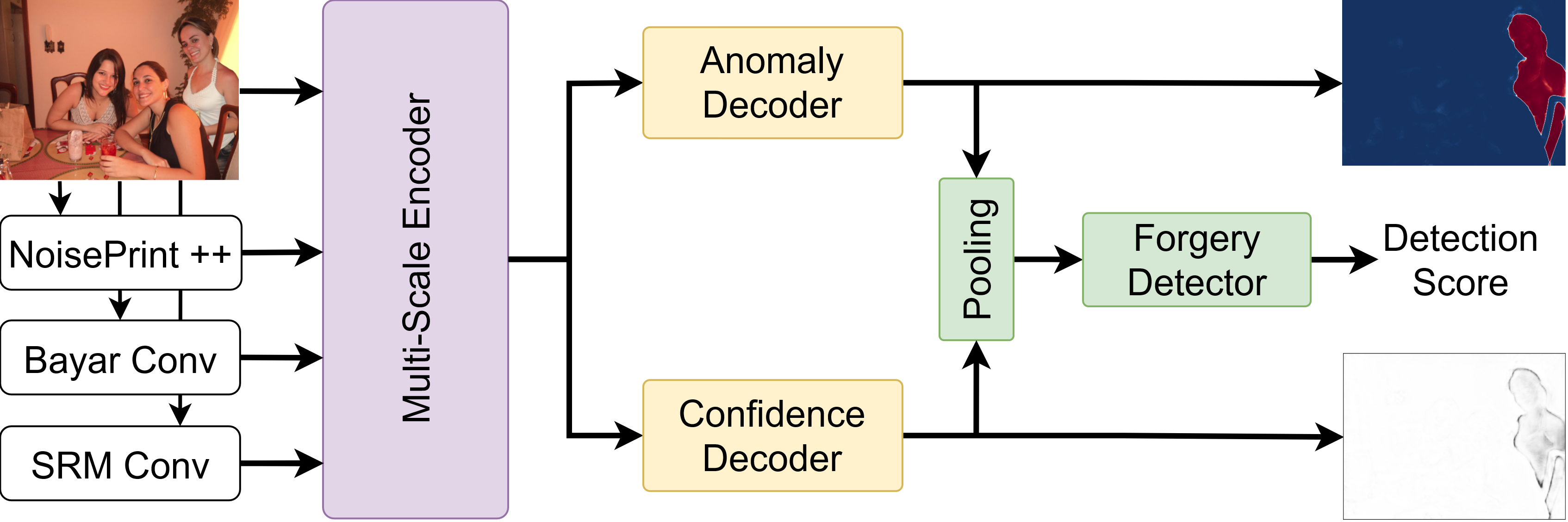}
\caption{Overview of the MMFusion Encoder-Decoder architecture for image localization and detection with multiple forensic filters. The RGB image and the output of each filter are fed into a Multi-Scale encoder, whose output is passed onto both the anomaly decoder, which produces a localization map, and the confidence decoder, which produces a confidence map. The two maps are then combined through a pooling module and passed into the forgery detector to produce the manipulation detection score.} \label{full}
\end{figure*}

Image forgery localization and detection are tasks the media forensics field has been working on for many years. Early works typically focused on a specific type of manipulation such as splicing\cite{mfcn}, copy-move\cite{cozzolino2014copy} or removal/inpainting\cite{inpaint}. More recently, deep-learning-based solutions of increasing robustness are proposed that are able to recognize multiple different types of manipulations\cite{mvss, pscc, trufor, mantranet, catnet2, crcnn, span, wang2022objectformer}. In order to be able to perform manipulation localization in a semantic-agnostic manner, these models need to suppress image content to reveal forensic artifacts. Most approaches achieve this by applying a high-pass filter to extract noise maps \cite{wang2022objectformer, span, mantranet, crcnn, mvss}. The most popular high-pass filters used are the ones proposed in the Steganalysis Rich Model (SRM)\cite{srm}, utilized in a wide variety of works\cite{rgbn, ermpc, span, mantranet, luo2021generalizing}, while the Bayar convolution\cite{bayar} is also used in a multitude of approaches\cite{mantranet, mvss, span, crcnn} and NoisePrint is used in a more recent model\cite{trufor}.

We hypothesize that those different forensic filters actually produce artifacts of complementary forensic capabilities. NoisePrint\cite{noiseprint} and its successor NoisePrint++\cite{trufor} produce artifacts that relate to camera model and editing history, thus displaying limited performance for copy-move manipulations (see Sec. \ref{ablation} for results supporting this statement). On the other hand, SRM\cite{srm} filters can identify edges and boundaries without relying on camera or compression/editing artifacts. These filters are, however, fixed by design (not trainable or changeable), a property that makes them vulnerable to adversarial attacks; whereas the Bayar convolution\cite{bayar}, on the other hand, learns the manipulation traces directly from data, proving more robust against malicious attacks. In this work we explore ways to expand existing state-of-the-art Image Manipulation Localization and Detection (IMLD) approaches to support multiple forensic filters as inputs. We start with TruFor\cite{trufor} as our baseline and propose utilizing NoisePrint++, SRM, and Bayar convolution as inputs auxiliary to the RGB image. We initially assess two different approaches to feature encoding: a late-fusion paradigm that extracts and encodes features from each modality (filter) separately, and an early-fusion paradigm that mixes the multi-modal features by early convolutional blocks. We also improve the decoder architecture of \cite{trufor} by introducing feature re-weighting (in both the Anomaly and Confidence decoders), improving the model's capability for recognizing and  localizing anomalies. We then propose the MMFusion architecture (Fig. \ref{full}), which employs early fusion and the aforementioned decoder architecture, and we also explore ways of explaining the predictions and understanding the unique capabilities of the different forensic filters of MMFusion. We find that the different filters excel at recognizing different kinds of image manipulations, validating our original hypothesis regarding their complementarity.

Furthermore, we investigate the capabilities of IMLD models without a special temporal-aware architecture for detecting and localizing manipulations in videos. We demonstrate that by using frame-level predictions with our MMFusion IMLD model we can reach state-of-the-art performance in Video Manipulation Localization and Detection (VMLD) tasks, documenting the merits of MMFusion while also highlighting the possibly limited complexity of the existing VMLD benchmarks.

A preliminary version of this work, with a simpler decoder architecture (without the proposed Feature Re-weighting Decoder) and without the studies on the explainability of multimodal IMLD models and the applicability of IMLD models to VMLD tasks, was presented in \cite{triaridis2024exploring}. Our main contributions in this paper are summarized as follows:
\begin{itemize}
    \item We compare the efficacy of different forensic filters, namely SRM, Bayar convolution and NoisePrint++, as inputs for deep networks performing forgery localization.
    \item We assess two distinct approaches for combining the outputs of different forensic filters for the purpose of image manipulation localization and detection, and we propose the MMFusion architecture.
    \item We propose, as part of the MMFusion architecture, a new Feature Re-weighing Decoder (FRD) that significantly increases localization performance.
    \item We also investigate the applicability of IMLD models, and our model specifically, on video datasets and compare it to state-of-the-art Video Manipulation Localization and Detection models. We compare our performance both with models specifically designed to tackle temporal inconsistencies for  VMLD and with simple IMLD models, and we provide a new baseline for the application of IMLD models on video without architectural changes.
    \item We propose a method for explaining the predictions of our multi-modal IMLD model by quantifying the contribution of each forensic filter for a specific image. This enables us to investigate the efficacy of each filter for different manipulation types and thus provide a deeper understanding of their predictive capabilities.
\end{itemize}

\section{Related Work}
\subsection{Image Manipulation Localization and Detection}
\label{related}
Traditional image forensics methods, e.g.  \cite{forensicssurvey,digitalforensics,compressionforensics}, have largely focused on detecting inconsistencies in low-level semantic-agnostic compression and internal camera artifacts. These artifacts can usually be revealed through high-pass filtering techniques that produce a noise-sensitive visualization of the image.

In recent times, various filters for noise extraction have been integrated into deep learning models to address the challenge of image manipulation localization and detection. Zhou et al. proposed RGB-N \cite{rgbn}, a two-stream Faster R-CNN network \cite{ren2015faster} that utilizes the RGB channel features as well as the extracted SRM-based noise features of the input image to detect visual inconsistencies and identify mismatches between authentic and tampered regions, originating from image splicing, to perform forgery detection with bounding boxes. In a similar fashion, Yang et al. showcased Constrained R-CNN (CR-CNN) \cite{crcnn}, a coarse-to-fine end-to-end architecture, which uses a learnable manipulation feature extractor (LMFE) based on a Bayar convolution, to create a unified feature representation for various manipulation types directly from the data. CR-CNN then follows two distinct stages: stage 1 performs manipulation technique classification and coarse manipulated region localization using the attention regional proposal network (RPN-A), while stage 2 fuses low- and high-level information to refine the global manipulation features. The model finally combines these refined features with the coarse localization information to further learn the finer local features and perform tampered region segmentation. Wu et al. \cite{mantranet} conducted experiments with various backbone network architectures and feature choices for proposing ManTraNet, a VGG-based manipulation localization and detection model, that integrates both SRM filters and Bayar convolution and is composed of three stages: adaptation, which adapts the manipulation trace feature for the anomaly detection task; anomalous feature extraction, which is inspired by human thinking and extracts anomalous features; and decision, which holistically considers anomalous features and classifies each pixel as either forged or not. Hu et al. presented SPAN (Spatial Pyramid Attention Network) \cite{span}, a framework for detecting and localizing various image manipulations by utilizing a hierarchical pyramid structure that models and encodes the relationships and the spatial positions of image patches at multiple scales using local self-attention blocks and position projection. It includes three blocks: a feature extractor, a spatial pyramid attention block, and a decision module applied on top of the output from the spatial pyramid attention propagation module to predict the localization mask. Moreover, Chen et al. introduced MVSSNet \cite{mvss}, a network that fuses the features from a ResNet-based edge-supervised branch (with a Sobel layer for edge enhancement) and a noise-sensitive branch using Bayar convolution via a trainable Dual Attention (DA) module in a late fusion paradigm, trained through multi-scale supervision. To combat the limited generalizability of MantraNet and SPAN, due to them being unable to fully take advantage of the spatial correlation, Liu et al. \cite{pscc} proposed PSCC-Net, which employs a Spatio-Channel Correlation Module (SCCM) that leverages the Gaussian function, to capture spatial and channel-wise correlations. PSCC-Net is a two-path structure architecture that follows a top-down path for extracting local and global features and a bottom-up path for detecting manipulations and estimating manipulation masks at multiple scales. It then uses dense cross-connections to fuse features across scales in a coarse-to-fine manner. Extending on their previous work that only targeted splicing forgery, Kwon et al. \cite{catnet2} expanded their research to additionally deal with copy-move forgery and introduced CatNetv2, which captures image acquisition camera-specific artifacts in the RGB domain and compression artifacts in the DCT domain and localizes the manipulated regions by considering the domains jointly. Similarly, Wang et al. defined ObjectFormer \cite{wang2022objectformer}, an end-to-end multi-modal framework that combines RGB features and frequency features and consists of a High-frequency Feature Extraction Module, an object encoder that uses learnable object queries to learn whether mid-level representations in images are coherent, and a patch decoder that produces refined global representations for manipulation detection and localization. Shi et al. \cite{dcnn} also proposed a dual domain-based CNN architecture with a spatial-domain CNN model (Sub-SCNN), that utilizes SRM filters and performs hierarchical feature extraction, and a frequency domain-based CNN model (Sub-FCNN) that extracts statistical features using the 3-level Daubechies-based Discrete Wavelet Transformation (DWT). Song et al. \cite{tbpnet} created the Tri-Path Backbone Architecture (TPB-Net) that consists of three DenseNet169 networks in a feature pyramid structure, to integrate features from different levels. They introduce a Dual-path Compressed Sensing Attention (DCSA) module to facilitate feature fusion, with the reasoning that high-level feature maps generally contain richer semantic information. Focusing specifically on inpainting region localization, Daryani et al. \cite{irlnet} defined a CNN-based deep learning model called IRL-Net, which includes three main modules: the enhancement module that tries to enhance inpainting traces with a Bayar layer, the encoder which includes four residual units to avoid vanishing/exploding gradients, and a Decoder that uses attention to map the learned high-level features extracted by the encoder. Finally, Guillaro et al. leveraged NoisePrint \cite{noiseprint}, a noise extractor proposed by Cozzolino et al. that is trained in a self-supervised manner to extract camera-specific artifacts and expanded its use in TruFor \cite{trufor}, where it is used jointly with RGB images in a dual-branch CMX \cite{cmx} architecture.

Contrary to most of the above works, that rely on a single forensic filter, our approach innovatively explores strategies for combining the outputs of three diverse noise extractors, leveraging their complementary capabilities to develop a robust end-to-end image forgery detection and localization model.

\subsection{Video Manipulation Localization and Detection}

Early works on video forensics, much like image forensics, relied heavily on non-learning based signal processing techniques to extract forensic artifacts. These methods were generally restricted in the types of forgeries they could detect and in the accuracy of said detection \cite{video_survey}.

In recent years, the trend has shifted toward deep learning-based approaches, enabling the development of networks capable of recognizing a broad range of forgeries. However, most methods have focused on the detection of a narrow set of very specific forgeries \cite{zampoglou2019detecting}, such as frame insertion and deletion \cite{gironi2014video, long2017c3d, shanableh2013detection}, or most commonly face deepfakes \cite{yang2019exposing, luo2021generalizing, jiang2020deeperforensics, yan2023ucf}. Very few approaches have tackled the general problem of video forgery detection directly. The advent of models like VideoFACT \cite{videofact, video_survey} illustrates the shift toward deep learning-based frameworks. VideoFACT \cite{videofact} incorporates forensic and contextual embeddings to capture traces left by manipulation and check for variations in forensic traces introduced by video coding. Subsequently, to estimate the quality and the relative importance of these local embeddings, it employs a deep self-attention mechanism.

Similarly to most works on IMLD, the few existing VMLD methods do not leverage multiple forensic filters and their potential complementarity.

\section{Methodology}
\label{sec:methodolgy}

\subsection{Encoder-Decoder Architecture}

\begin{figure*}[ht]
\centering
\includegraphics[width=0.6\textwidth]{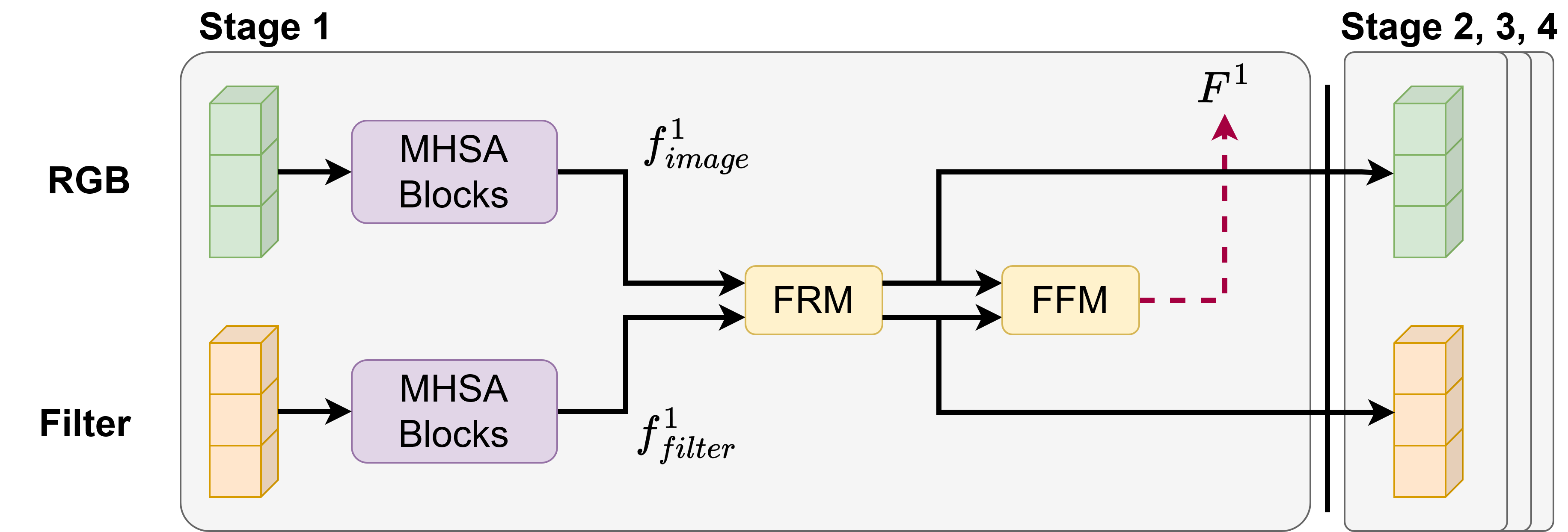}
\caption{Architecture of the dual-branch encoder of \cite{cmx}. The encoder is made of 4 stages of Multi-Head Self Attention (MHSA) blocks to produce feature maps $f_{mod}^i$ for modality $mod\in\{image, filter\}$ and stage $i\in\{1,2,3,4\}$. These are then fused and rectified by the FRM and FFM modules to produce the outputs $F^i$ at each scale $i$. The feature map set $F=\{F^i, i=1,...4\}$ is the final output returned by the encoder.} \label{encoder}
\end{figure*}

Our goal is to extend an existing encoder-decoder-based architecture to be able to use multiple forensic filters (SRM\cite{srm}, Bayar convolution\cite{bayar}, NoisePrint++\cite{trufor}) in tandem, so as to produce more robust representations for the IMLD task. To this end we adopt the general architecture of TruFor \cite{trufor}, i.e., as illustrated in Fig. \ref{full} we use an encoder, an anomaly decoder, a confidence decoder, and a forgery detector; and we follow TruFor's two-phase training regime for anomaly localization and detection, respectively. The encoder follows the popular dual-branch architecture proposed in \cite{cmx} and illustrated in Fig. \ref{encoder} for a single forensic filter, comprising of 4 stages of Multi-Head Self Attention (MHSA) blocks\cite{segformer} that produce feature maps $f_{mod}^i$ of different scales: $\frac{H}{2^{i+1}}\times\frac{W}{2^{i+1}}\times C_i$, where $i\in\{1,2,3,4\}, mod\in\{image, filter\}$, $H$ and $W$ are the spatial dimensions of the input image and $C_i$ is the channel dimension of the output at stage (and scale) $i$. The two MHSA blocks' outputs (for the RGB image and the filter) in each stage are rectified through a Cross-Modal Feature Rectification Module (FRM)\cite{cmx} that exploits the interactions between the two input modalities (RGB and NoisePrint++ in the case of TruFor). The FRM uses features from both modalities to produce weighted channel- and spatial-wise feature maps that are residually added for both modalities to perform channel- and spatial-wise rectification. The two sets of feature maps are then combined using a Feature Fusion Module (FFM)\cite{cmx}, whose outputs $F^i$ (having the same dimensions as the outputs of the MHSA blocks at each scale $i$) for $i\in\{1,2,3,4\}$ collectively constitute the encoder output $F=\{F^i, i=1,...4\}$ (see Table \ref{table:notationsummary} for notation summary, and references to the relevant architecture diagrams). The FFM consists of an information exchange stage, where a cross-attention mechanism exchanges information between modalities and produces two sets of mixed feature maps, and a fusion stage where the feature maps are merged into a single output through a residual Multilayer Perceptron (MLP) module that uses $1\times1$ convolutions. The Decoders illustrated in Fig. \ref{full} are, in the case of TruFor \cite{trufor}, MLP-based decoders proposed in \cite{segformer}. 

\begin{table}[t]
\caption{Main symbols used in Sec. \ref{sec:methodolgy} for denoting feature maps.}
\begin{tabular}{lp{0.74\columnwidth}}
    \hline
    Symbol         & Description\\
    \hline
    $f^{i}_{mod}$   & Feature maps returned by the MHSA block of the encoder for modality $mod\in\{image, filter\}$ at stage $i$ (Fig. \ref{encoder})\\
    $F^i$           & Fused feature maps returned by the FFM module of the encoder at stage $i$ (Fig. \ref{encoder}, \ref{early_conv}) \\
    $f^{i}_{filter}$& Feature maps returned by the MHSA block of the late fusion encoder for filter $filter\in\{noiseprint, srm, bayar\}$ at stage $i$ (Fig. \ref{late_fusion})\\
    $F^{i}_{filter}$& Feature maps returned by the FFM module of the late fusion encoder for filter $filter\in\{noiseprint, srm, bayar\}$ at stage $i$ (Fig. \ref{late_fusion})\\
    $f_a$           & Mixed features produced by the early fusion module (Fig. \ref{early_conv})\\
    $F=\{F^i\}$     & Output feature maps of the encoder (Fig. \ref{dec})\\
    $D=\{D^i\}$      & Re-weighted feature maps fed to the MLP-based decoder (Fig. \ref{dec})\\
    \hline
\end{tabular}
\label{table:notationsummary}
\end{table}

Utilizing this architecture one can combine RGB images with an auxiliary forensic modality to perform Image Manipulation Localization. In \cite{trufor} Guillaro et al. use their own feature extractor NoisePrint++, however a multitude of other forensic filters' outputs, such as Bayar convolution\cite{bayar} or SRM\cite{srm}, can be utilized. These filters are analyzed in Sec. \ref{filters}. We assess two different ways of extending the encoder architecture to multiple auxiliary modal inputs: a late fusion paradigm, where each auxiliary modality is combined with RGB inputs separately using a dual-branch architecture \cite{cmx} (Sec. \ref{s:late_fusion}), and an early fusion paradigm where auxiliary modalities are combined early before being utilized as input to the dual-branch encoder together with the RGB inputs (Sec. \ref{s:early_fusion}). We then propose the MMFusion architecture, that extends the encoder architecture to multiple auxiliary modal inputs using early fusion and introduces a feature re-weighting step in the decoders.

\begin{figure*}[ht]
\centering
\includegraphics[width=0.7\textwidth]{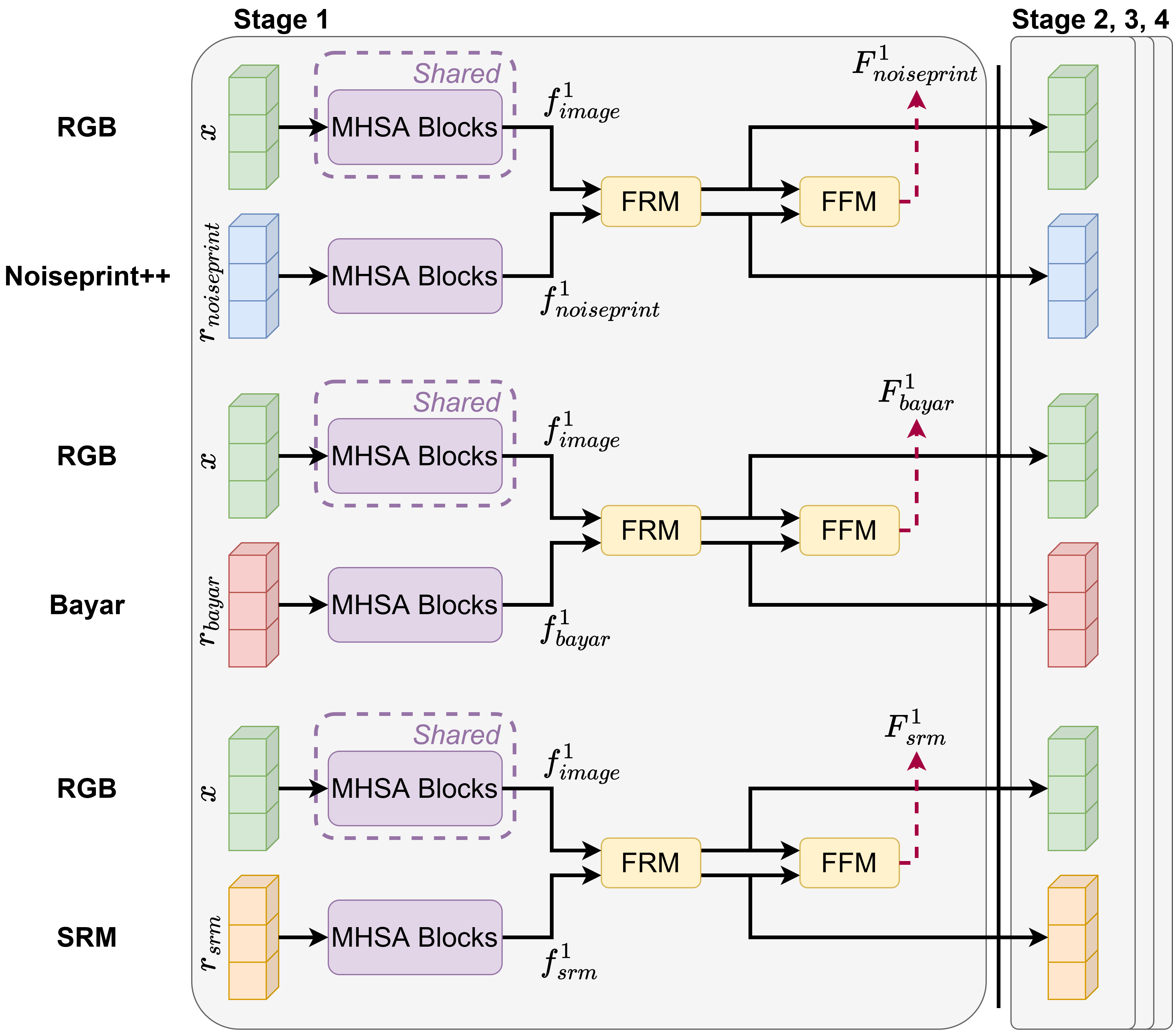}
\caption{Proposed architecture of the encoder for fusion of multiple forensic filters by late fusion with weight sharing. The filters' outputs and the RGB image are fed into separate MultiHead Self-Attention (MHSA) blocks of the dual-branch CMX encoder, with the outputs rectified and combined by the FRM and FFM modules to produce the feature maps. These are propagated through different stages to create feature maps of varying scales. The weights of the MHSA blocks of all RGB branches are shared to increase regularization.} \label{late_fusion}
\end{figure*}

\subsection{Auxiliary (a.k.a. filter) modalities}
For both early- and late-fusion approaches, we use the outputs of three forensic filters, namely NoisePrint++, SRM and Bayar convolution, as inputs that are auxiliary to the RGB image. We choose these filters as they are widely used in the relevant literature (Sec. \ref{related}), they showcase good performance and also complement each other well: SRM is a static feature extractor that mostly extracts edge features, while NoisePrint++ and Bayar convolution are trainable modules that are, however, trained with different objectives. NoisePrint++ is trained in a self-supervised contrastive manner as a camera ``fingerprint'' extractor, while Bayar convolution is directly trained for IMLD in a supervised setting, as explained below.
\label{filters}
\subsubsection{NoisePrint++}
In \cite{noiseprint} Cozzolino et al. proposed Noiseprint, a CNN-based model designed to extract camera-model-based artifacts from RGB images while suppressing image content. In \cite{trufor} they expanded their approach, namely NoisePrint++, to be able to recognize and extract artifacts related to the editing history of an image (e.g. compression, resizing, gamma correction). NoisePrint++ is trained in a supervised contrastive manner\cite{supcon}: a batch of images is provided, from which patches are extracted from different locations. Then the patches go through different editing pipelines. Patches extracted from the same source image, the same location, and with the same editing history are considered positive samples, while others are considered negative. In our work we use NoisePrint++ as a pretrained feature extractor.
\subsubsection{SRM}
Another way to suppress the image content and highlight forensic traces and noise is through static high-pass filters, the most common of which are the ones proposed for producing residual maps for the Steganalysis Rich Model (SRM) \cite{srm}. Out of the 30 high-pass filters proposed in \cite{srm}, we use the 3 most commonly used in the literature, e.g. as in \cite{rgbn, mantranet, span}, which are displayed in Fig. 4 of \cite{rgbn}. Following  \cite{srm} and \cite{rgbn}, the outputs of these three filters are truncated and combined to form the final noise descriptor, referred in the sequel as SRM filter.
\subsubsection{Bayar Convolution}
In contrast to using static high-pass filters for noise extraction, Bayar et al.\cite{bayar} proposed the constrained convolutional layer as a noise extractor that adaptively learns manipulation traces from data. We use the constrained convolutional layer as an extra noise feature extractor and refer to it as Bayar convolution. For both our multi-modal fusion approaches the Bayar convolutional layer is pretrained alone in a dual branch CMX encoder \cite{cmx}
(as also done in our ablation study for examining the effect of using Bayar as the sole auxiliary input alongside the RGB image; see Sec. \ref{ablation} and the results for ``CMX (Bayar)'' in Table \ref{table:ablation}) and then used with its weights frozen.

\begin{figure*}[ht]
\centering
\includegraphics[width=0.9\textwidth]{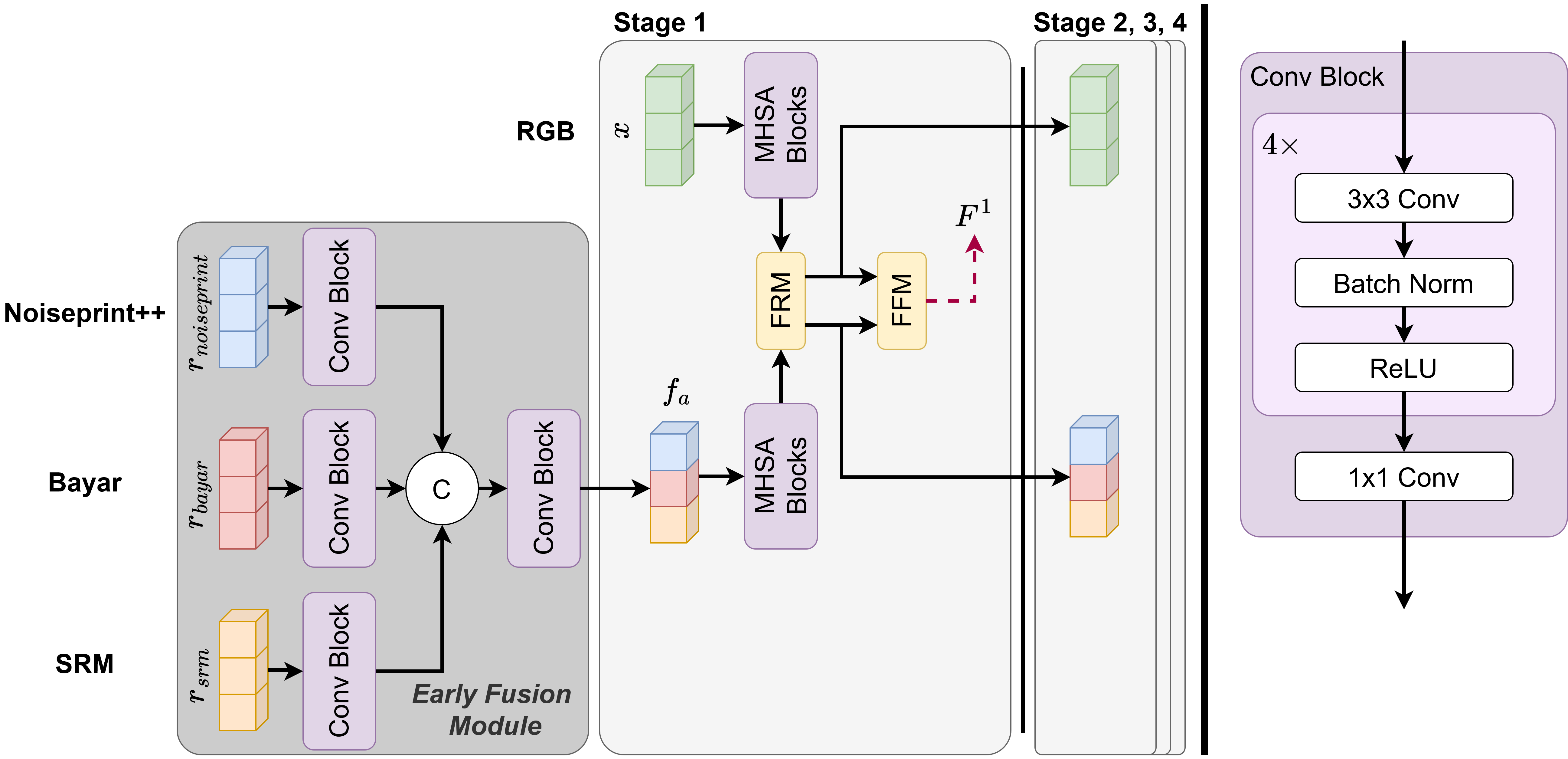}
\caption{Proposed architecture of the encoder for fusion of multiple forensic filters by early convolutions. On the left, we illustrate the structure of the encoder. More specifically, the filters' outputs are initially fused by early convolutional blocks in the Early Fusion Module, to produce the mixed features $f_a$. These features and the RGB image are then fed into separate MultiHead Self-Attention (MHSA) blocks of a dual-branch CMX encoder, with the outputs rectified and combined by the FRM and FFM modules to produce the feature maps. These are propagated through different stages to create feature maps of varying scales. The structure of the convolutional block is presented on the right side.}\label{early_conv}
\end{figure*}

\begin{figure*}[ht]
\centering
\includegraphics[width=0.8\textwidth]{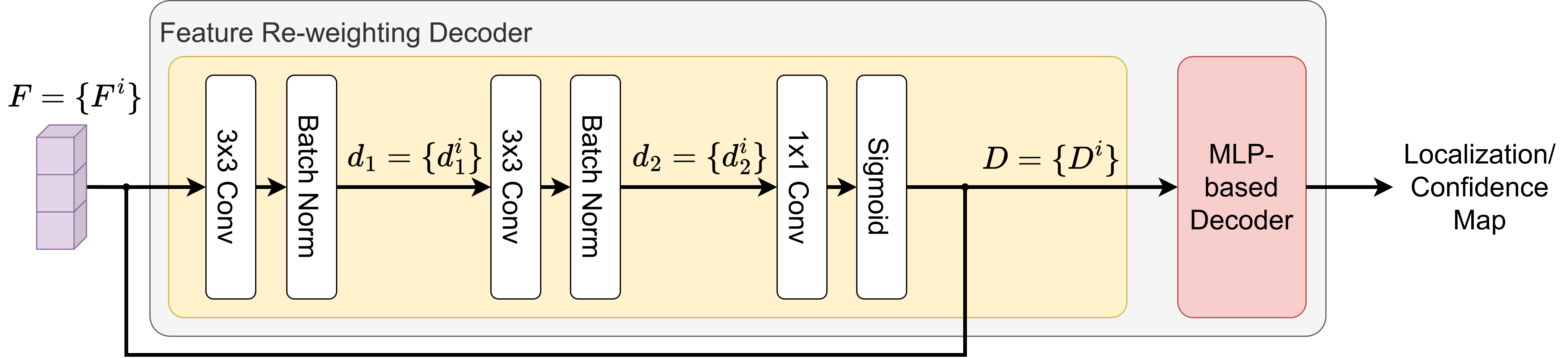}
\caption{Proposed architecture of the Feature Re-weighting Decoder (FRD). The feature maps $F$ returned from the encoder are processed through convolutional layers, batch normalization and activation functions, and weighted channel- and spatial-wise feature maps that enhance subtle variations in the input maps are produced. These are then passed to the MLP-based decoder to generate the localization/confidence map.} \label{dec}
\end{figure*}

\subsection{Late Fusion}
\label{s:late_fusion}
For the late fusion approach we extract the auxiliary representations $r_{noiseprint}$, $r_{srm}$, $r_{bayar}$ of the RGB image $x$ from the NoisePrint++, SRM and Bayar filters respectively. Then the output of each filter is fed together with the original RGB input into a dual-branch CMX encoder $\mathcal{E}$, made of 4 Multi-Head Self Attention (MHSA) blocks \cite{segformer} that produce 4-scale feature maps $f_{mod}^i$, where $mod\in\{image, filter\}$, $filter\in\{noiseprint, srm, bayar\}$ and $i\in\{1,2,3,4\}$. These feature maps are passed to and rectified through the FRM and FFM modules, to produce the final features of the encoder $F^i_{filter} = \mathcal{E}_{filter}(x, r_{filter})$ as shown in Fig. \ref{late_fusion}. 

The outputs $F^i_{filter}$ of the three encoders for a given $i$ are concatenated, and the resulting set of feature maps for $i\in\{1,2,3,4\}$ constitutes the final output $F$ of the encoder, which is then passed to the decoders (as illustrated in Fig. \ref{full}). In this late fusion approach we use the same decoder architecture as in TruFor for both the anomaly and confidence decoders. Like other multi-modal approaches, this approach is prone to overfitting and the ``modality imbalance'' problem \cite{mmHARD, mmProto}, where different modalities converge and overfit at different rates, thus hindering joint optimization. To tackle this we make the weights of the modules along the RGB branch shared across all 3 encoders to increase regularization. We also employ Dropout before the anomaly decoder as the complete encoder is rather large and the simple MLP-based decoder is prone to overfitting.

\subsection{Early Fusion}
\label{s:early_fusion}

For the early fusion approach we extract again the same auxiliary representations $r_{noiseprint}$, $r_{srm}$, $r_{bayar}$ of the RGB image $x$.
The inputs are then passed through our novel Early Fusion Module $\mathcal{EFM}$ to produce the mixed features $f_a = \mathcal{EFM}(r_{noiseprint}, r_{srm}, r_{bayar})$ as shown in Fig. \ref{early_conv}. The $\mathcal{EFM}$ consists of 3 independent convolutional blocks, one for each auxiliary modality, and one final convolutional block that performs feature mixing. The convolutional blocks are good at early visual processing, resulting in a more stable optimization\cite{ec}, thus aiding in mixing the features from different modalities smoothly. The mixed features $f_{a}$ and RGB image $x$ are used as input for a dual-branch CMX encoder\cite{cmx}, in the same manner as in TruFor. This is a particularly lightweight approach to expanding the TruFor architecture to handle multiple auxiliary modalities, as it does not significantly increase the number of parameters of the network (68.9M params compared to TruFor's 68.7M, as reported in \cite{trufor}).

Each of the convolutional blocks mentioned above consists of four $3\times3$ convolutions followed by a $1\times1$ convolutional layer to resize the output to 3 channels (Fig. \ref{early_conv}). There is a batch normalization (BN) and a ReLU layer after each $3\times3$ convolutional layer. The output channels for the $3\times3$ convolutional layers are [24, 48, 96, 192].

\subsection{Feature Re-weighting Decoder (FRD)}

For the decoder part of our network we enhance the MLP-based decoder used in CMX \cite{cmx} to implement both the anomaly and the confidence decoder (see Fig. \ref{full}), with feature re-weighing. The motivation behind this is that, for harder-to-recognize manipulations, the encoder may produce feature maps that are not dissimilar enough between the image's original and manipulated regions. Thus we employ feature re-weighting to enhance the differences between these parts. The process works by taking the feature maps $F^i$ of the feature map set $F$ produced by the encoder and re-weighing them (resulting in re-weighted feature maps $D^i$), before feeding the multi-layer feature map set $D=\{D^i, i=1,...4\}$ into an MLP-based decoder as shown in Fig. \ref{dec}. The latter produces the final decoder output. This re-weighting ensures that the subtle variations become more pronounced, making it easier for the subsequent MLP-based decoder to differentiate and accurately process the manipulated content. This significantly enhances the network's ability to detect difficult-to-recognize manipulations. The re-weighted feature maps $D=\{D^i, i=1,...4\}$ are calculated (Fig. \ref{dec}) as follows:
\begin{align*}
    d_1^i &= BN(Conv_{3\times3}(F^i))  \\
    d_2^i &= BN(Conv_{3\times3}(d_1^i))  \\
    D^i  &=  sigmoid(Conv_{1\times1}(d_2^i)) * F^i
\end{align*}
where $BN$ denotes Batch Normalization and $Conv_{3\times3}$ and $Conv_{1\times1}$ are convolutions with ${3\times3}$ and ${1\times1}$ kernels, respectively.

\section{Experiments}
\subsection{Experimental Setup}
\subsubsection{Training} We follow the training procedure proposed by Guillaro et al. \cite{trufor}: first, we jointly train the encoder and anomaly decoder; after that, we train the confidence decoder and the forgery detector, while the encoder and anomaly decoder are kept frozen. For both training phases in IMLD we use the datasets used by Kwon et al.\cite{catnet2}, and sample an equal number of images from each one for every epoch. For the VLMD task, we use the datasets proposed in \cite{videofact}. The employed training datasets are summarized in Table \ref{table:data}.

\subsubsection{Testing} 
\label{sec:testdatasets}
For testing, we evaluate our model on five IMLD datasets: Coverage\cite{coverage}, Columbia\cite{columbia}, Casiav1+\footnote{Casiav1+ is a modification of the Casiav1 dataset proposed by Chen et al.\cite{mvss} that replaces authentic images that also exist in Casiav2 with images from the COREL\cite{corel} dataset to avoid data contamination.}\cite{casia} and DSO-1\cite{dso1}, which are widely used in the relevant literature, and CocoGlide \cite{trufor}. The latter is a diffusion-based manipulation dataset proposed recently by Guillaro et al \cite{trufor}, that uses the COCO validation dataset, \cite{microsoftcoco} an object mask with its corresponding label as the forgery region and the text prompt, and feeds them to GLIDE \cite{nichol2022glide} to generate new synthetic objects. As for the other aforementioned datasets, Coverage accommodates copy-move forgery manipulations, while Columbia and DSO-1 focus on splicing manipulations. Finally, Casiav1+ includes a variety of image manipulations such as splicing, copy-move, and removal.

We also evaluate our models on VCMS, VPVM and VPIM \cite{videofact}, which are datasets for Video Manipulation Localization and Detection. VCMS contains splicing manipulations, while VPVM and VPIM contain manipulations made with standard video editing operations (blurring, gamma adjustment etc), utilizing different strengths that make them visually perceptible (in VPVM) or imperceptible (in VPIM), respectively. For all video datasets, we sample just the first frame of each video to evaluate our method (hence the number of frames in Table \ref{table:test_data} is 300 for each of the manipulated / non-manipulated classes, equal to the number of test videos for each class that we retrieved from \footnote{https://huggingface.co/datasets/ductai199x/video-std-manip}). We did such a radical frame sampling because in early experiments we observed that using just one frame per video yielded the same results as averaging across all video's frames ($\pm 0.01$), while naturally requiring significantly less runtime.

The employed testing datasets are summarized in Table \ref{table:test_data}.

\begin{table}[t]
\centering
\begin{tabular}{r c c c}
 & & \multicolumn{2}{c}{Number of Images}\\
 \cline{3-4}
Dataset      & & Real & Manipulated\\
\hline
Casiav2\cite{casia} & &7.491& 5.105\\
IMD2020\cite{imd2020} && 414& 2.010\\
FantasticReality\cite{fantasticreality} && 16.592&19.423\\
cm\_coco\cite{catnet2}&&- & 200.000\\
bcm\_coco\cite{catnet2}&&-  &200.000\\
bcmc\_coco\cite{catnet2}&&-  &200.000\\
sp\_coco\cite{catnet2}&& -&200.000\\
\hline
VCMS\cite{videofact}&& 48.000 & 48.000\\
VPVM\cite{videofact}&& 48.000 & 48.000\\
VPIM\cite{videofact}&& 48.000 & 48.000\\
\hline
\end{tabular}
\caption{Number of real and manipulated images (or video frames) in each training dataset. (Image datasets above line, video datasets below line)}
\label{table:data}
\end{table}

\begin{table}[t]
\centering
\begin{tabular}{r c c c}
 & & \multicolumn{2}{c}{Number of Images}\\
 \cline{3-4}
Dataset      & & Real & Manipulated\\
\hline
Coverage\cite{coverage} & &100& 100\\
Columbia\cite{columbia} && 183& 180\\
Casiav1+\cite{casia} && 800&921\\
DSO-1\cite{dso1}&&100  &100\\
CocoGlide\cite{trufor}&&512 &512\\
\hline
VCMS\cite{videofact}&& 300 & 300\\
VPVM\cite{videofact}&& 300 & 300\\
VPIM\cite{videofact}&& 300 & 300\\
\hline
\end{tabular}
\caption{Number of real and manipulated images (or video frames) in each test dataset. (Image datasets above line, video datasets below line)}
\label{table:test_data}
\end{table}

\subsubsection{Evaluation Measures} For localization performance we follow most previous works, e.g. \cite{trufor,rgbn,crcnn,mantranet,span,mvss,pscc,catnet2,wang2022objectformer,tbpnet,irlnet,dcnn}, and report average pixel-level performance using the F1 measure, which uses the ground truth and the prediction mask to determine the True Positives (TP), True Negatives (TN), False Positives (FP) and False Negatives (FN). The F1 and the inverse F1 scores are then computed and the maximum value between the two is returned. TP are the pixels where the ground truth and the prediction mask overlap to correctly identify manipulated regions, TN are the pixels where the ground truth and the prediction mask overlap to correctly identify non-manipulated regions, FP are the pixels where the prediction incorrectly identifies manipulated regions not present in the ground truth and FN are the pixels where the prediction mask fails to identify manipulated regions present in the ground truth. We use a fixed threshold of 0.5, where pixels with higher value indicate the predicted manipulated regions, as setting a best threshold per test dataset \cite{catnet2} or even per image\cite{trufor}, like some other previous works have done, is not realistic in practical scenarios where the ground truth is not available, thus leading in exaggerated performance estimates. For detection, similarly to e.g. \cite{trufor,rgbn,crcnn,mantranet,span,mvss,pscc,wang2022objectformer,tbpnet}, we calculate the image-level Area Under Curve (AUC), which is a measure that evaluates the model’s ability to separate the two classes (manipulated and or not) over various thresholds. AUC takes values in the range [0.5,1.0], where a perfect model would score 1.0, while a randomly guessing one would score 0.5. We also employ balanced accuracy (bAcc) as in \cite{trufor}, which is the arithmetic mean of sensitivity and specificity, with the threshold set once again to 0.5.
\subsubsection{Implementation} All models are implemented in PyTorch and trained on a single consumer-grade NVIDIA GPU (either an RTX 4090 or an RTX 3090), using an effective batch size of 24 for 100 epochs. Physical batch size ranged from 4 to 8 depending on the model and an effective batch size of 24 was reached by utilizing gradient accumulation. We use a Dropout rate of 0.3 for the methods proposed in this paper. The MHSA modules were initialized with ImageNet-pretrained weights as proposed in \cite{cmx,cmnext}. We utilized an SGD optimizer with an initial learning rate of 0.005, momentum of 0.9, weight decay of 0.0005 and a polynomial learning rate schedule. For training augmentations we followed the protocol of Guillaro et al.\cite{trufor}: resized the images in the [0.5-1.5] range, performed random cropping of size 512$\times$512 and JPEG compression with a random Quality Factor QF$\in$[30,100].

\subsection{Evaluation and Comparisons on Image Manipulation Datasets}
We compare our methods with recent state-of-the-art approaches for IMLD. Following Guillaro et al. we consider methods with open source models provided and we exclude models that use part of our testing datasets for training to avoid bias. Overall, we compare with TruFor\cite{trufor}, CAT-Netv2\cite{catnet2}, ManTraNet\cite{mantranet}, PSCC-Net\cite{pscc}, SPAN\cite{span}, Constrained R-CNN\cite{crcnn}, MVSS-Net\cite{mvss}. Results are presented in Table \ref{table:localization}.

\begin{table*}[t]

\centering
 \caption{Average pixel-level F1 scores for the localization task on the considered models and image datasets. Best (higher) scores in bold and second best scores underlined. Results for all models except for the proposed ones are taken from \cite{trufor}.}
 \label{table:localization}
\begin{tabular}{ | c|| c  c  c  c  c | c | }
 \hline
 Model & Coverage & Columbia & Casiav1+ & CocoGlide & DSO-1 & AVG \\
 \hline \hline

TruFor\cite{trufor}      & 0.600      & 0.859      & 0.737      & 0.523      & \textbf{0.930}  & 0.729\\
CAT-Netv2\cite{catnet2}   & 0.381      & 0.859      & 0.752      & 0.434      & 0.584  & 0.602\\
ManTraNet\cite{mantranet}   & 0.317      & 0.508      & 0.180      & 0.516      & 0.412  & 0.387\\
PSCC-Net\cite{pscc}    & 0.473      & 0.604      & 0.520      & 0.515      & 0.458  & 0.514\\
SPAN\cite{span}        & 0.235      & 0.759      & 0.112      & 0.298      & 0.233  & 0.327\\
CR-CNN\cite{crcnn}      & 0.391      & 0.631      & 0.481      & 0.447      & 0.289  & 0.448\\
MVSS-Net\cite{mvss}    & 0.514      & 0.729      & 0.528      & 0.486      & 0.358  & 0.523\\
\hline
Late Fusion (Sec. \ref{s:late_fusion}) & 0.641     & 0.864      & 0.775      & \underline{0.574}      & \underline{0.899}  & \underline{0.751}\\
Early Fusion (Sec. \ref{s:early_fusion}) & \underline{0.663}      & \textbf{0.888}      & \underline{0.784}      & 0.553      & 0.863  &  0.750\\
MMFusion (Early Fusion with FRD) & \textbf{0.700}      & \underline{0.876}      & \textbf{0.794}      & \textbf{0.591}      & 0.866  &  \textbf{0.765}\\
\hline
\end{tabular}

\end{table*}

\begin{table*}[h]

\centering
 \caption{Results of the Area Under Curve (AUC) and balanced accuracy (bAcc) measures for the detection task on the considered models and image datasets. Best (higher) scores in bold and second best scores underlined.}
 \label{table:detection}
\begin{tabular}{|c||c c c c c c c c c c |c c|}
\hline
& \multicolumn{2}{c}{Coverage}  & \multicolumn{2}{c}{Columbia}  & \multicolumn{2}{c}{Casiav1+}  & \multicolumn{2}{c}{CocoGlide} & \multicolumn{2}{c|}{DSO-1}    & \multicolumn{2}{c|}{AVG}\\
Model & AUC & bAcc & AUC & bAcc & AUC & bAcc & AUC & bAcc & AUC & bAcc & AUC & bAcc \\
\hline
\hline
TruFor\cite{trufor}      & 0.770& 0.680     & 0.996& \textbf{0.984}     & 0.916&   0.813   & 0.752& 0.639     & \textbf{0.984}&\underline{0.930}  & 0.884& \underline{0.809}\\
CAT-Netv2\cite{catnet2}  & 0.680& 0.635     & 0.977& 0.803     & \textbf{0.942}& 0.838      & 0.667&  0.580    & 0.747& 0.525 & 0.803& 0.676\\
ManTraNet\cite{mantranet}   & 0.760&  0.500    & 0.810&0.500      & 0.644&  0.500    & \textbf{0.778}&  0.500    & 0.874& 0.500 & 0.773&0.500\\
PSCC-Net\cite{pscc}    & 0.657&   0.473    & 0.300& 0.604   & 0.869&  0.520    & \underline{0.777}&  0.515    & 0.650& 0.458 & 0.651&0.514\\
SPAN\cite{span}        & 0.670&  0.235   & \textbf{0.999}& 0.759     & 0.480&0.112      & 0.475& 0.298     & 0.669&0.233  & 0.659&0.327\\
CR-CNN\cite{crcnn}      & 0.553& 0.391     & 0.755&  0.631    & 0.670& 0.481     & 0.589&0.447      & 0.576& 0.289 & 0.629&0.448\\
MVSS-Net\cite{mvss}    & 0.733& 0.514     & 0.984&0.729      & \underline{0.932}&  0.528    & 0.654& 0.117     & 0.552&0.358  & 0.771&0.449\\
\hline
Late Fusion (Sec. \ref{s:late_fusion}) & 0.792 & 0.720  & 0.977& 0.822     & 0.930&  \textbf{0.860}    & 0.760& \underline{0.677}     &0.958 & 0.830 & 0.884&0.782\\
Early Fusion (Sec. \ref{s:early_fusion}) & \textbf{0.839} & \textbf{0.770}     &0.996 & \underline{0.962}     & 0.929 &  \underline{0.845}    &0.755 &   0.660   & \underline{0.966}& \textbf{0.935} & \textbf{0.897} & \textbf{0.834}\\
MMFusion (Early Fusion with FRD) & \underline{0.837}& \underline{0.765} & \underline{0.998}&  0.814    & 0.931 &   \textbf{0.860}   &0.775 & \textbf{0.699}   & 0.923& 0.735 &\underline{0.893} & 0.776\\
\hline
\end{tabular}

\end{table*}

\begin{figure*}[h!]
\centering
\begin{tabular}{ccccc}
Image & Ground Truth & Late Fusion & Early Fusion & MMFusion \\

\includegraphics[width=.15\linewidth]{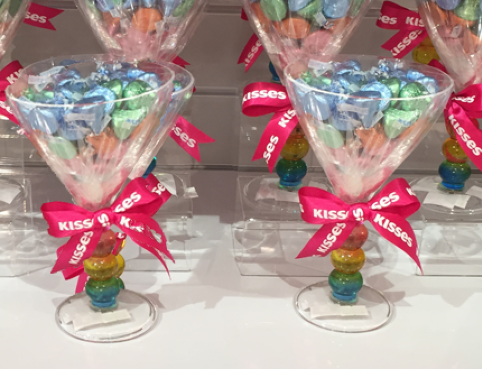} &
\frame{\includegraphics[width=.15\linewidth]{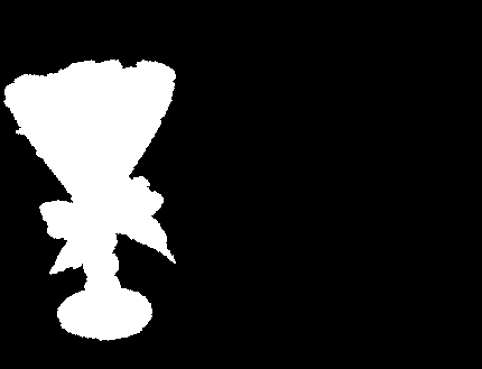}} &
\frame{\includegraphics[width=.15\linewidth]{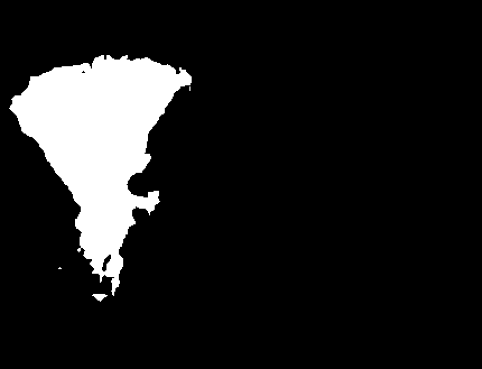}} &
\frame{\includegraphics[width=.15\linewidth]{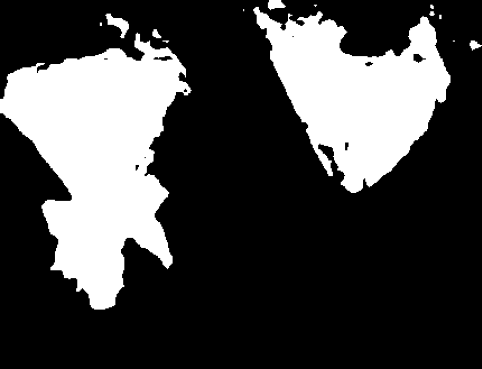}} &
\frame{\includegraphics[width=.15\linewidth]{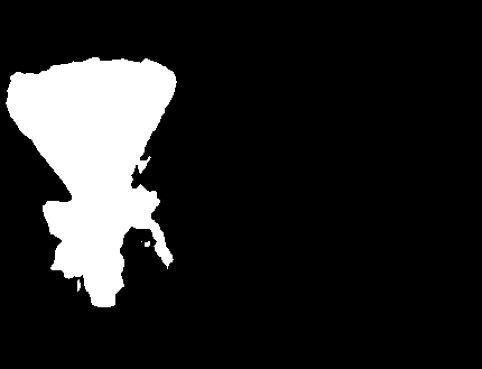}} \\

\includegraphics[width=.15\linewidth]{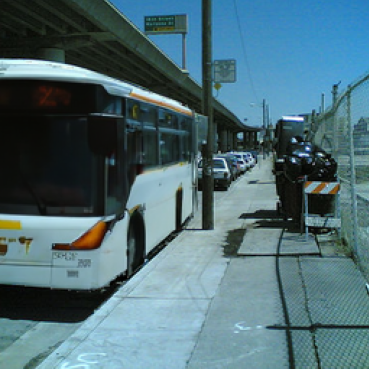} &
\frame{\includegraphics[width=.15\linewidth]{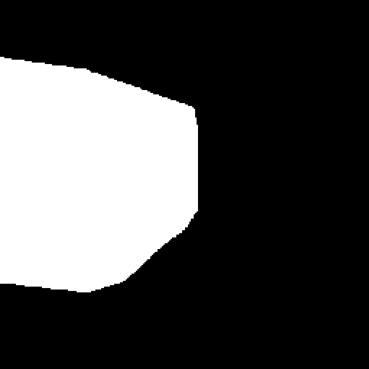}} &
\frame{\includegraphics[width=.15\linewidth]{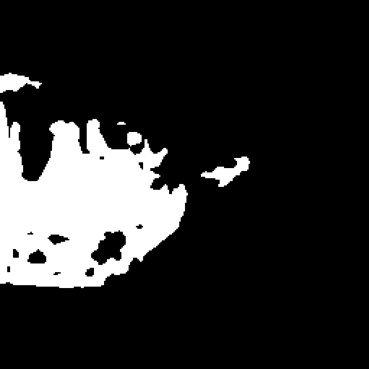}} &
\frame{\includegraphics[width=.15\linewidth]{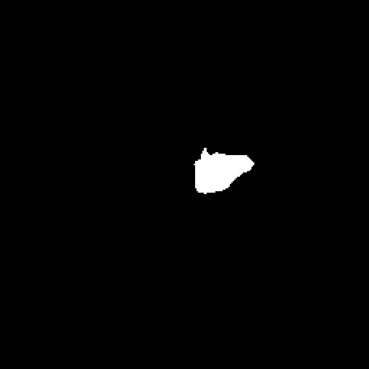}} &
\frame{\includegraphics[width=.15\linewidth]{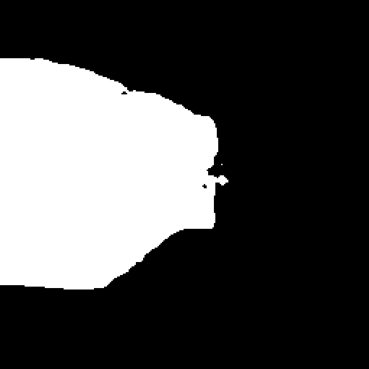}} \\

\includegraphics[width=.15\linewidth]{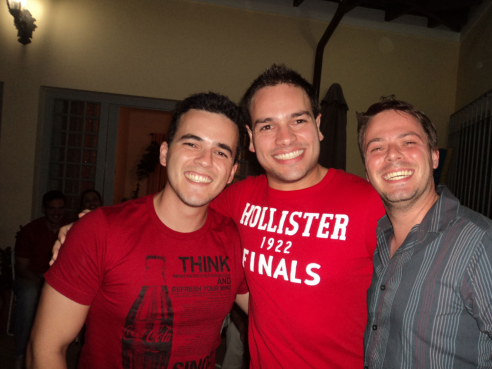} &
\frame{\includegraphics[width=.15\linewidth]{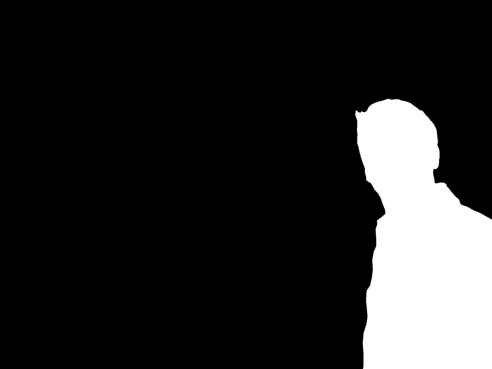}} &
\frame{\includegraphics[width=.15\linewidth]{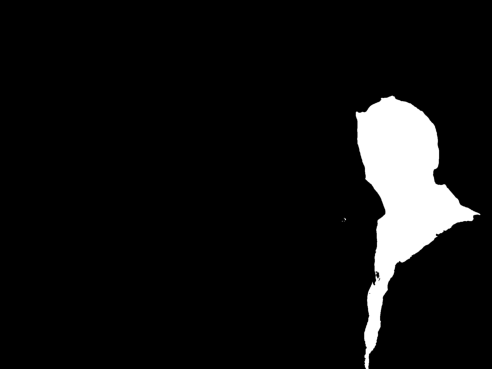}} &
\frame{\includegraphics[width=.15\linewidth]{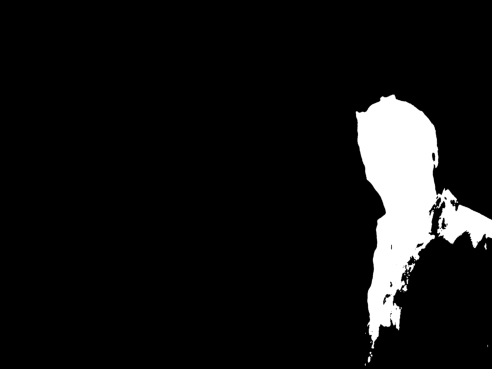}} &
\frame{\includegraphics[width=.15\linewidth]{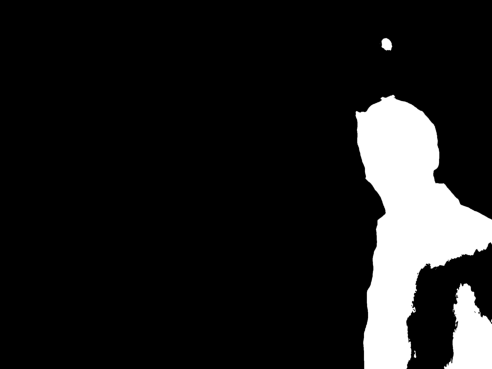}} \\

\end{tabular}
\caption{Qualitative results, showing for each image the Ground Truth mask of the manipulated region and the corresponding prediction of each of the examined / proposed filter fusion approaches. Source dataset of each image: top row: Coverage; middle row: CocoGlide; bottom row: DSO-1.}
\label{table:qual}
\end{figure*}

Both of our multi-modal fusion approaches, as well as the proposed MMFusion architecture, showcase state-of-the-art performance, being either the best or second-best model for every dataset. We also observe that the proposed decoder employed in MMFusion further increases the average F1 by 1.5\%, reaching a value of 76.5\%. Especially for the Coverage dataset that contains only copy-move forgeries, MMFsusion surpasses the previous best, TruFor, by 10\%. The only dataset where we do not achieve top performance is DSO-1, where our best approach (Late Fusion) is 3\% behind TruFor. 

We also compare across models in terms of detection performance and present the results in Table \ref{table:detection}.  Notably, our early fusion variant and the MMFusion architecture demonstrate exceptional performance, surpassing the state-of-the-art on average. Particularly noteworthy is the outstanding performance on the Coverage dataset, where they achieve a remarkable improvement of nearly 7\% in terms of the Area Under the Curve (AUC) and 9\% in terms of balanced accuracy (bAcc) compared to the prior leading method. Our late fusion approach also exhibits competitive AUC performance, but falls slightly behind the TruFor model in terms of bAcc. This disparity in bAcc performance could potentially be attributed to the size of our late fusion model, which makes it susceptible to overfitting.

Finally, the effectiveness of our proposed approaches is illustrated with qualitative results in Fig. \ref{table:qual}, where we see that all our fusion approaches can approximately localize the existing manipulation(s), with MMFusion achieving more accurate and complete localization. More specifically, in the first image (coming from the Coverage dataset), MMFusion most closely matches the Ground Truth, while the Early Fusion model without FRD over-predicts, highlighting the top of the original glass, and Late Fusion misses key areas like the ribbon. In the second image (from the CocoGlide dataset), MMFusion again performs the best, accurately capturing most of the manipulated region, while the other two fusion approaches leave gaps in the shape of the bus or even fail to detect it all together. For the third image (from the DSO-1 dataset), MMFusion provides the most accurate prediction of the three, as it correctly detects the lower right portion of the shirt as manipulated.

\subsection{Ablation Study}
\label{ablation}

\begin{table*}[h!]
\centering
\caption{Average pixel-level F1 scores for the localization task on the considered ablation study variants and datasets. Parameter count in Millions. Runtime in milliseconds for a single image of the Casiav1+ dataset on an RTX 3090 GPU. Best scores (higher for avg F1, lower for params/runtime) in bold and second best scores underlined. }
\resizebox{\textwidth}{!}{
\begin{tabular}{|c||c c c c c |c| c c |}
\hline
Version         & Coverage  & Columbia  & Casiav1+  & CocoGlide & DSO-1    & AVG& Params(M)&Runtime(ms)\\
\hline
\hline
CMX (NP++)  & 0.577      & 0.884      & 0.761      & 0.516      & \underline{0.895}  &  0.726& \underline{68.3}      & 34.9    \\
CMX (Bayar) & 0.592      & 0.872      & 0.774      & 0.566      & 0.776  &   0.716& \textbf{68.1}      & \underline{34.2}  \\
CMX (SRM)   & 0.630      & 0.834      & \underline{0.791}      & \underline{0.585}      & 0.792  & 0.726& \textbf{68.1}    & \textbf{34.0}    \\
\hline
\begin{tabular}[c]{cc}Late Fusion (Sec. \ref{s:late_fusion}) -\\No weight sharing\end{tabular}  & 0.611      & \textbf{0.912}      & 0.760      & 0.566      & 0.785  & 0.727& 200.7   & 79.1\\
Late Fusion (Sec. \ref{s:late_fusion}) & 0.641      & 0.864      & 0.775      & 0.574      & \textbf{0.899}  & \underline{0.751}& 152.3    & 77.2\\
Early Fusion (Sec. \ref{s:early_fusion}) & \underline{0.663}      & \underline{0.888}      & 0.784      & 0.553     & 0.863  &  0.750 & 68.9      & 42.0\\

MMFusion (Early Fusion with FRD) & \textbf{0.700}      & 0.876      & \textbf{0.794}      & \textbf{0.591}      & 0.866  &  \textbf{0.765} & 68.9 & 43.6 \\
\hline
\end{tabular}}
\\[1ex]

\label{table:ablation}
\end{table*}

\begin{figure*}[h!]
\centering
\begin{tabular}{cccccc}
Image &
Ground Truth & CMX(NP++) & CMX(Bayar) & CMX(SRM) & MMFusion \\

\includegraphics[width=.14\linewidth]{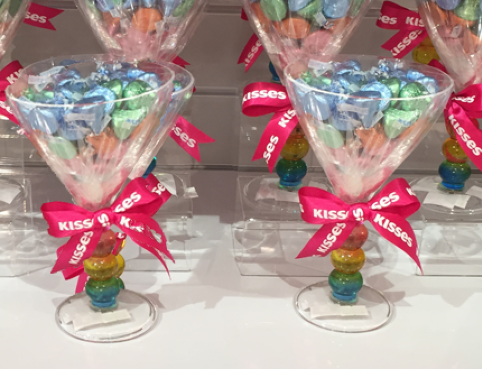} & \includegraphics[width=.14\linewidth]{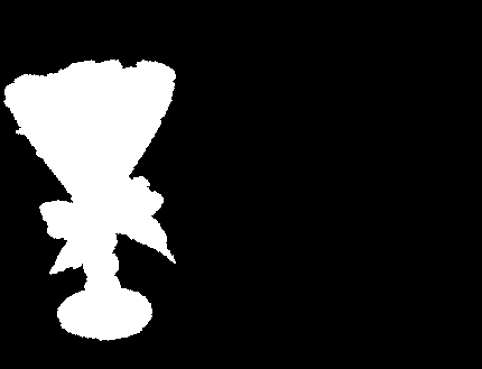} &
\includegraphics[width=.14\linewidth]{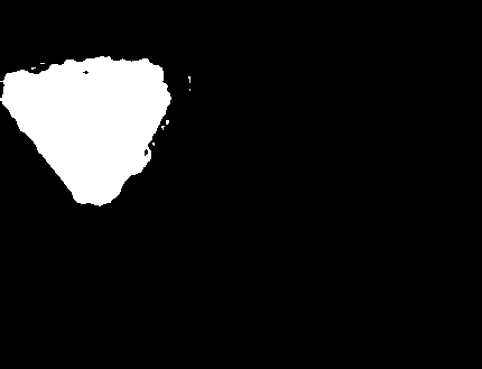} &
\includegraphics[width=.14\linewidth]{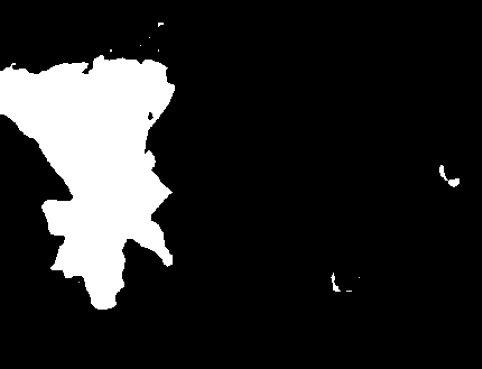} &
\includegraphics[width=.14\linewidth]{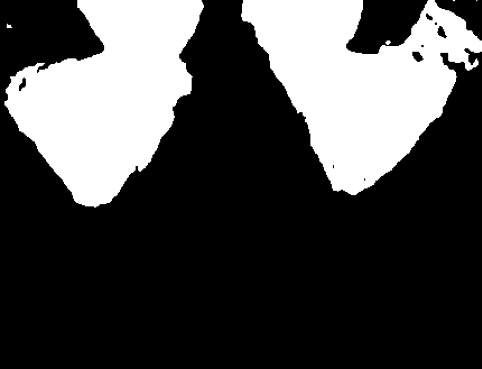} &
\includegraphics[width=.14\linewidth]{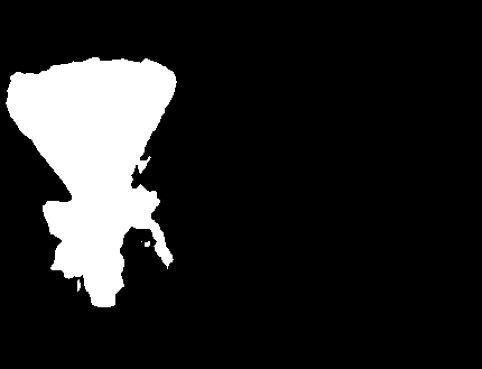} \\

\includegraphics[width=.14\linewidth]{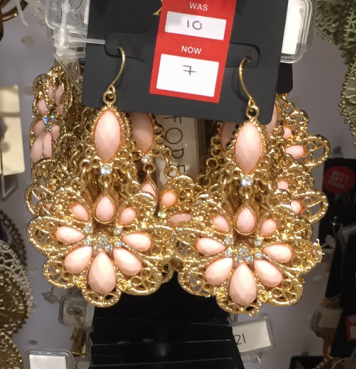} & \includegraphics[width=.14\linewidth]{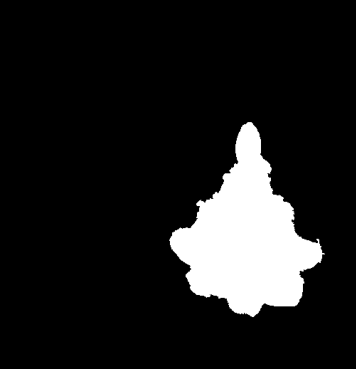} &
\includegraphics[width=.14\linewidth]{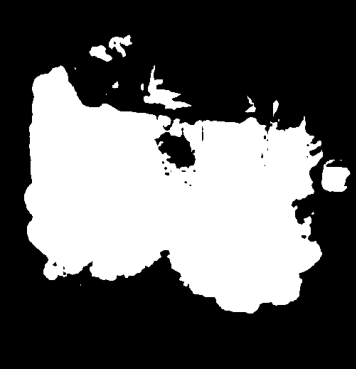} &
\includegraphics[width=.14\linewidth]{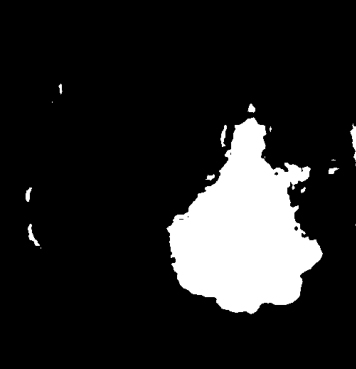} &
\includegraphics[width=.14\linewidth]{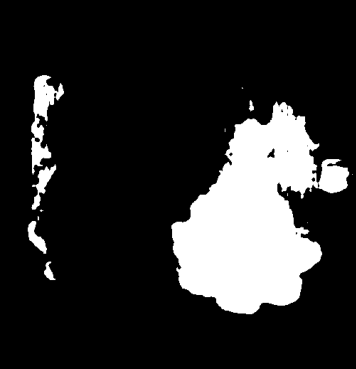} &
\includegraphics[width=.14\linewidth]{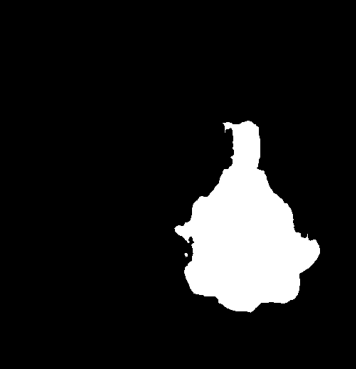} \\

\includegraphics[width=.14\linewidth]{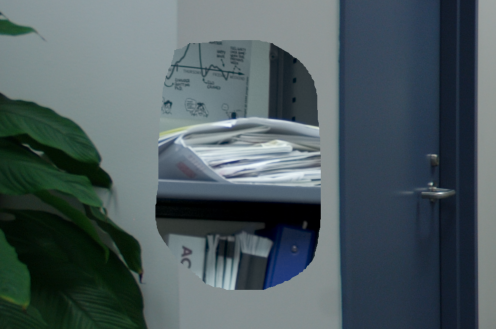} & \includegraphics[width=.14\linewidth]{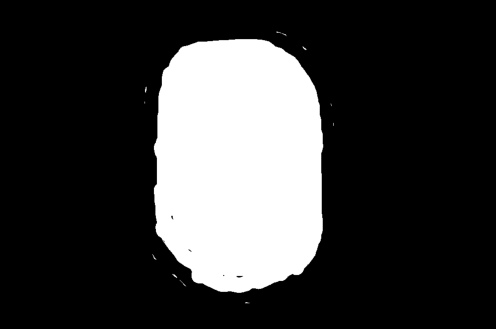} &
\includegraphics[width=.14\linewidth]{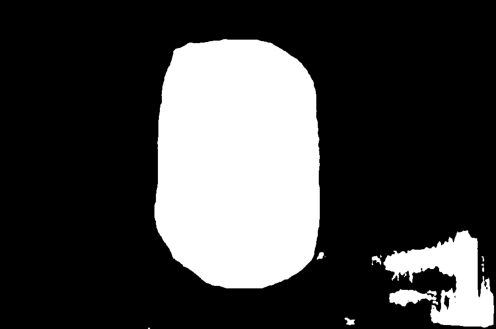} &
\includegraphics[width=.14\linewidth]{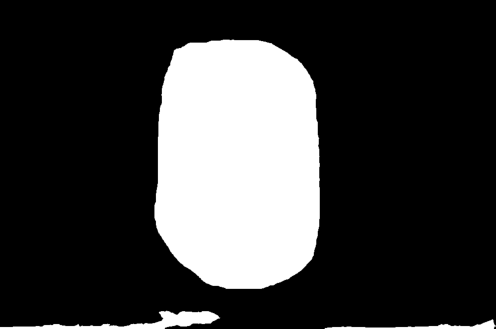} &
\includegraphics[width=.14\linewidth]{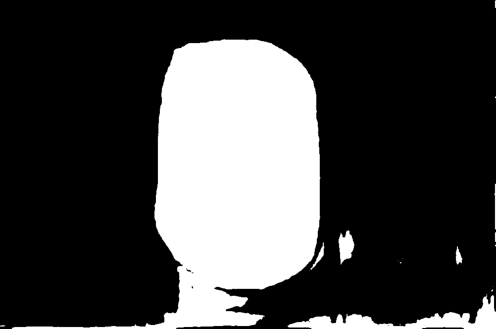} &
\includegraphics[width=.14\linewidth]{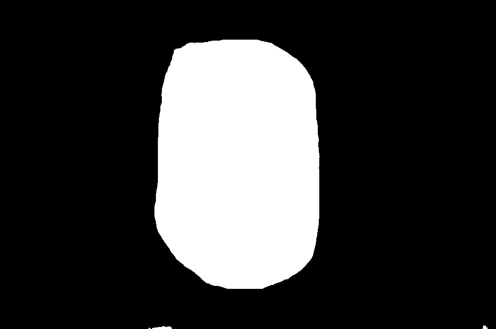} \\

\includegraphics[width=.14\linewidth]{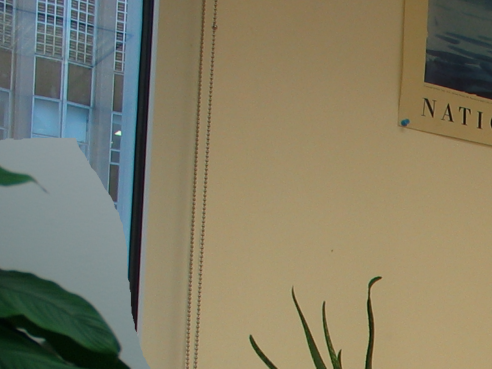} & \includegraphics[width=.14\linewidth]{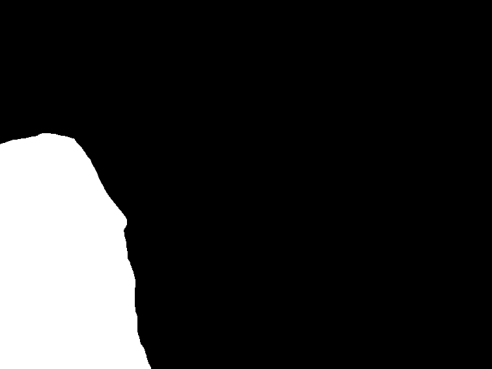} &
\includegraphics[width=.14\linewidth]{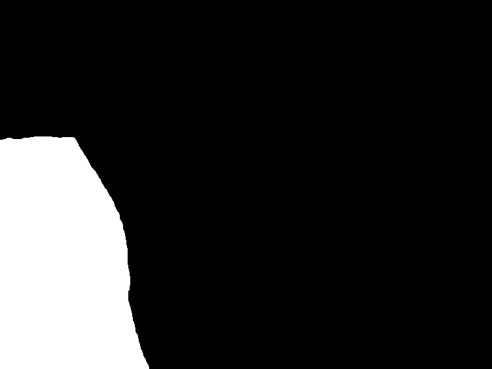} &
\includegraphics[width=.14\linewidth]{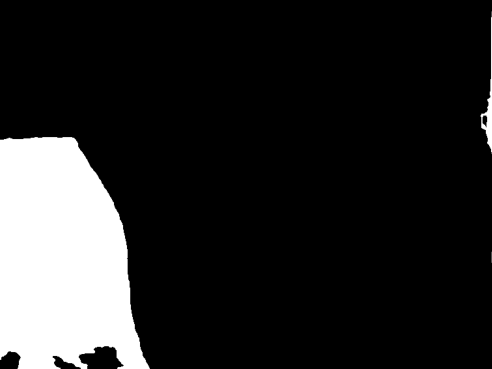} &
\includegraphics[width=.14\linewidth]{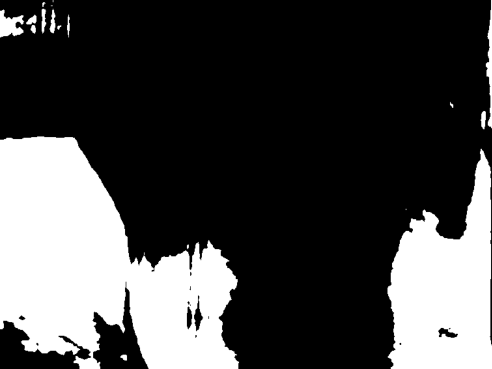} &
\includegraphics[width=.14\linewidth]{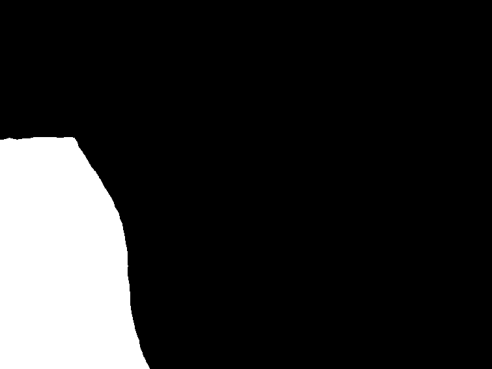} \\

\includegraphics[width=.14\linewidth]{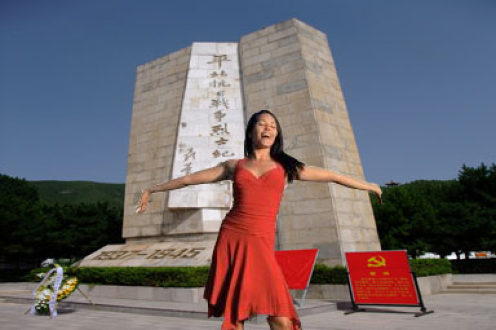} & \includegraphics[width=.14\linewidth]{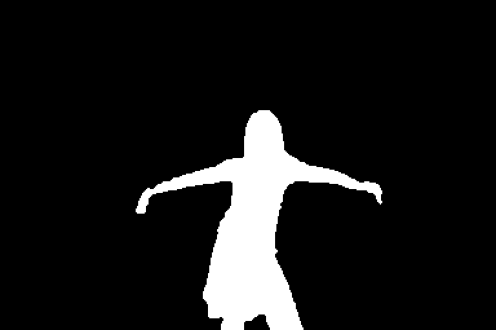} &
\includegraphics[width=.14\linewidth]{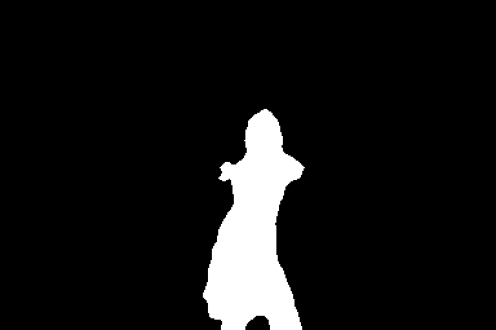} &
\includegraphics[width=.14\linewidth]{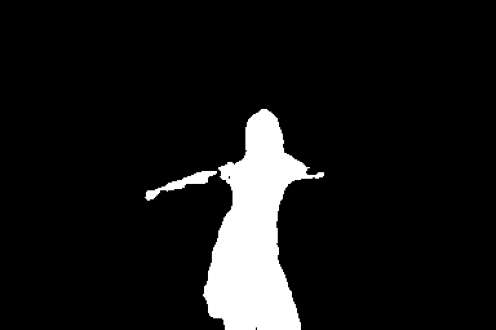} &
\includegraphics[width=.14\linewidth]{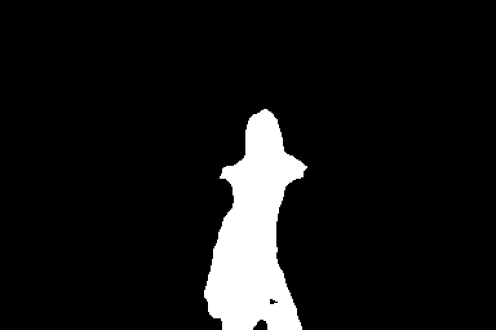} &
\includegraphics[width=.14\linewidth]{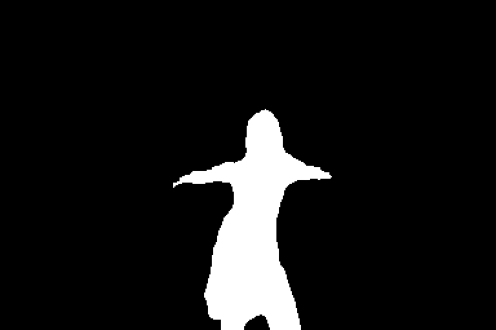} \\

\includegraphics[width=.14\linewidth]{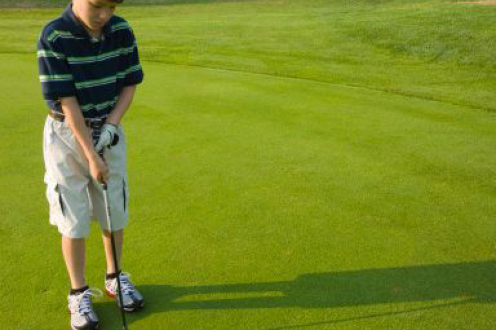} & \includegraphics[width=.14\linewidth]{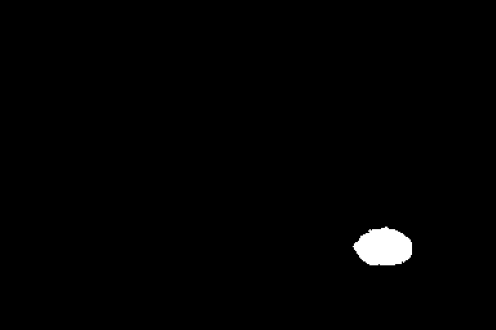} &
\includegraphics[width=.14\linewidth]{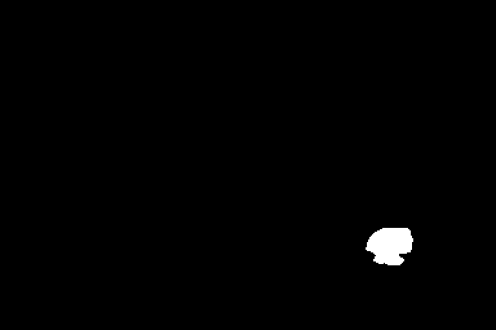} &
\includegraphics[width=.14\linewidth]{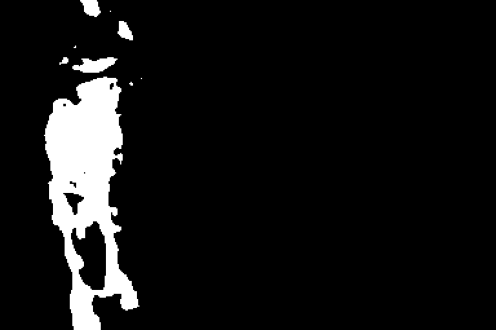} &
\includegraphics[width=.14\linewidth]{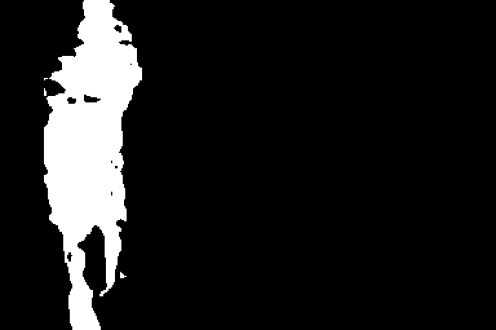} &
\includegraphics[width=.14\linewidth]{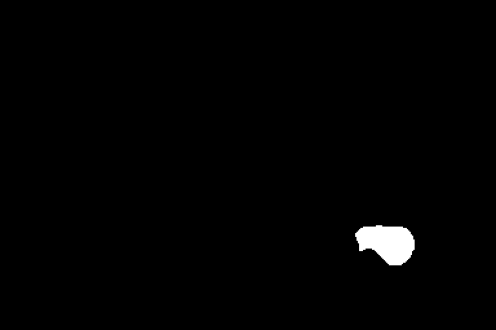} \\

\includegraphics[width=.14\linewidth]{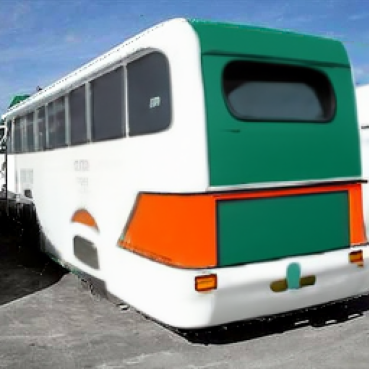} & \includegraphics[width=.14\linewidth]{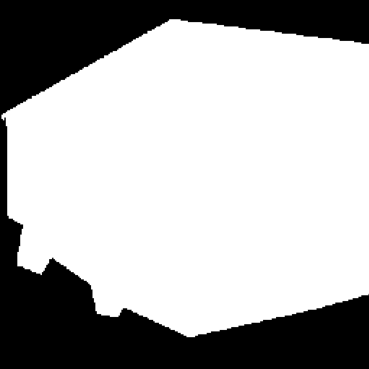} &
\includegraphics[width=.14\linewidth]{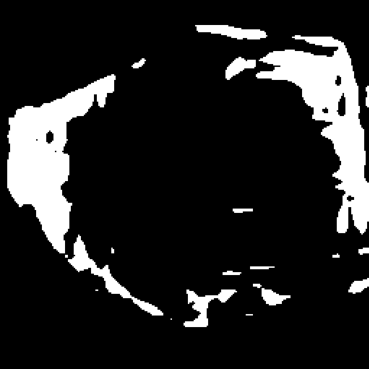} &
\includegraphics[width=.14\linewidth]{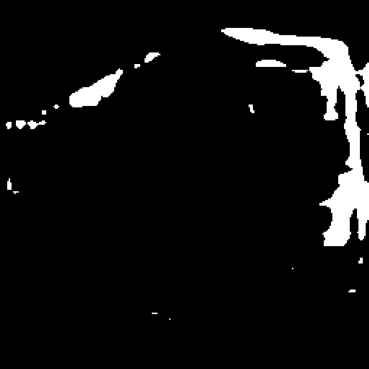} &
\includegraphics[width=.14\linewidth]{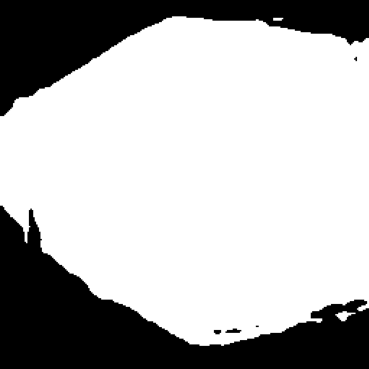} &
\includegraphics[width=.14\linewidth]{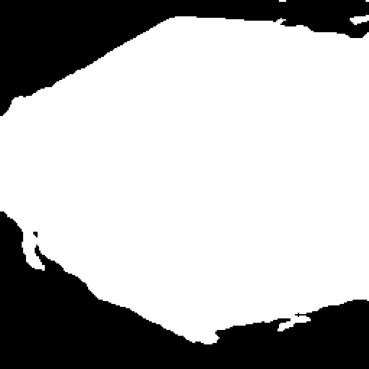} \\

\includegraphics[width=.14\linewidth]{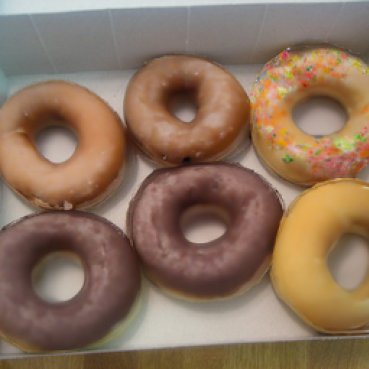} & \includegraphics[width=.14\linewidth]{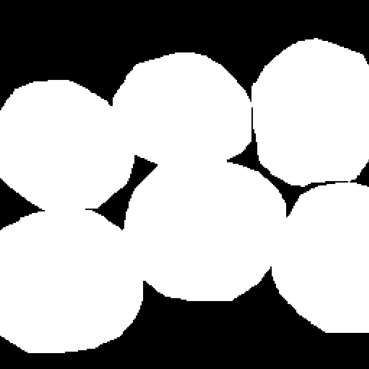} &
\includegraphics[width=.14\linewidth]{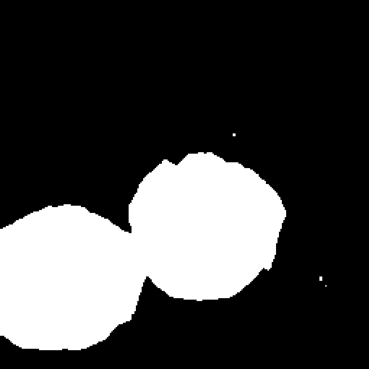} &
\includegraphics[width=.14\linewidth]{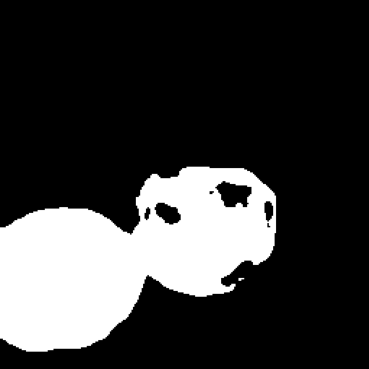} &
\includegraphics[width=.14\linewidth]{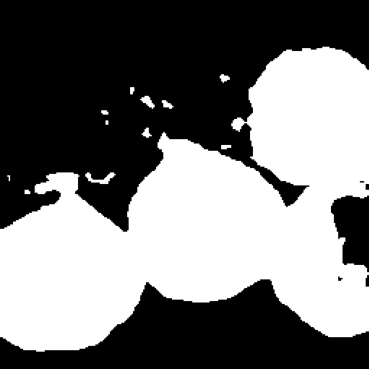} &
\includegraphics[width=.14\linewidth]{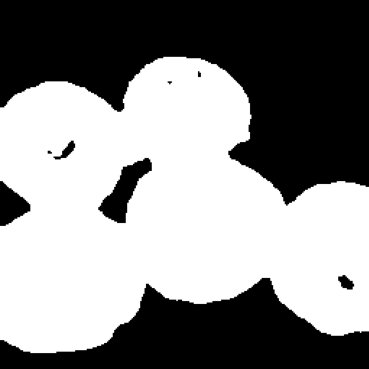} \\

\includegraphics[width=.14\linewidth]{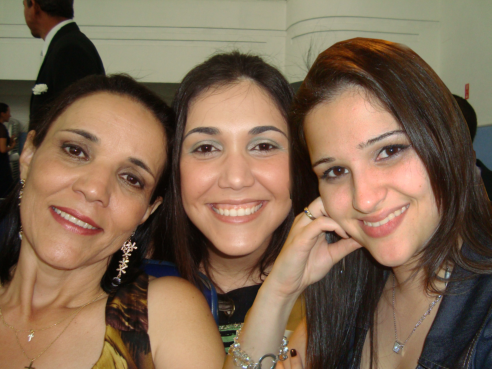} & \includegraphics[width=.14\linewidth]{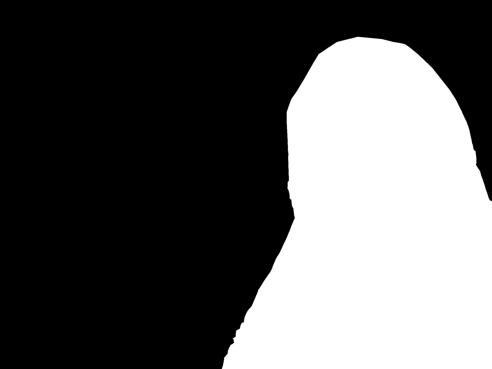} &
\includegraphics[width=.14\linewidth]{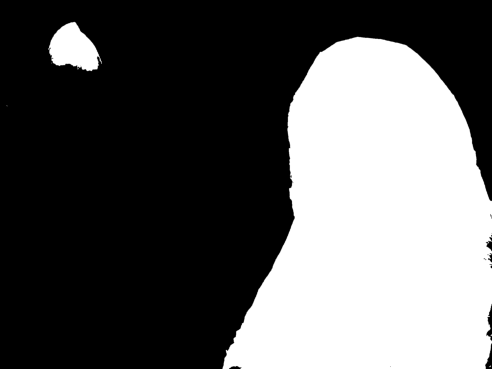} &
\includegraphics[width=.14\linewidth]{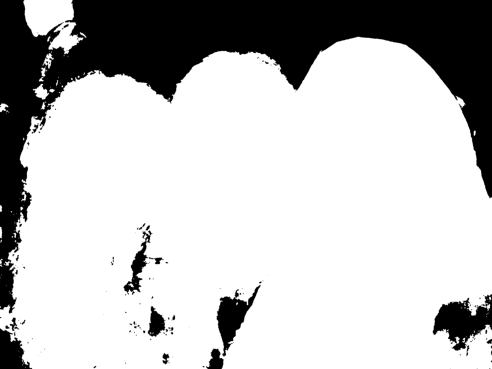} &
\includegraphics[width=.14\linewidth]{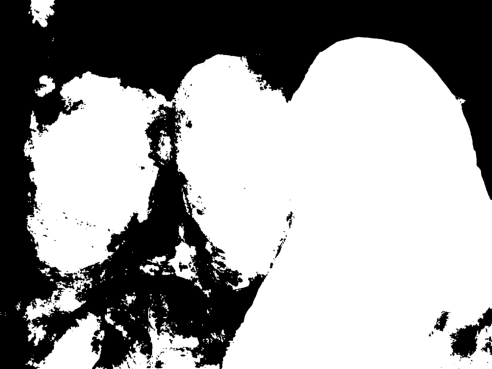} &
\includegraphics[width=.14\linewidth]{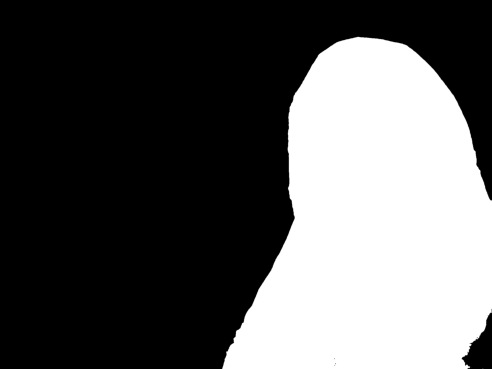} \\

\includegraphics[width=.14\linewidth]{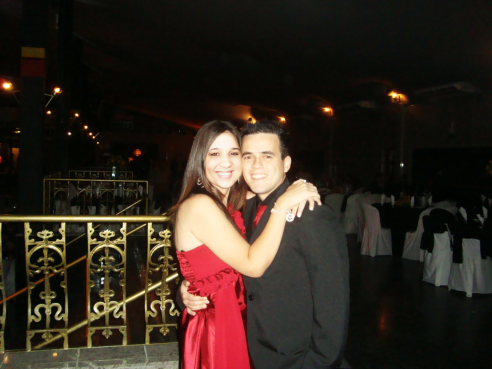} & \includegraphics[width=.14\linewidth]{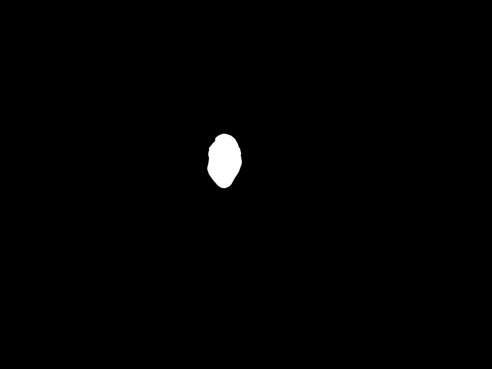} &
\includegraphics[width=.14\linewidth]{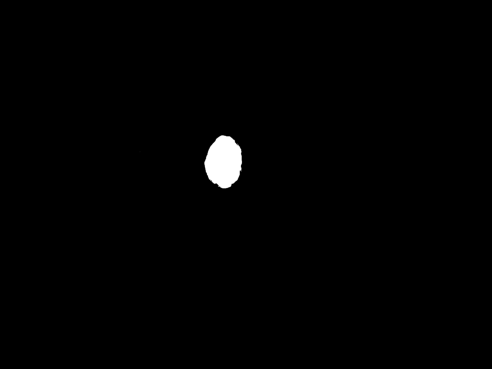} &
\includegraphics[width=.14\linewidth]{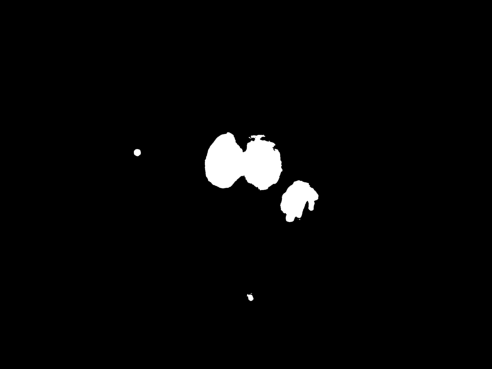} &
\includegraphics[width=.14\linewidth]{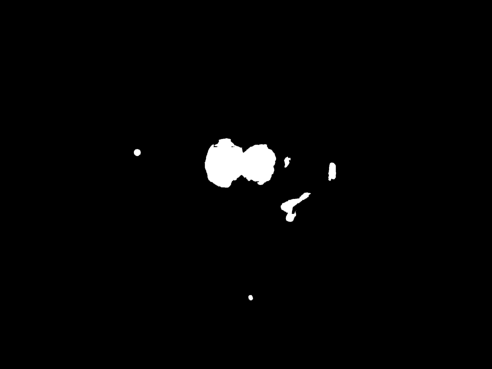} &
\includegraphics[width=.14\linewidth]{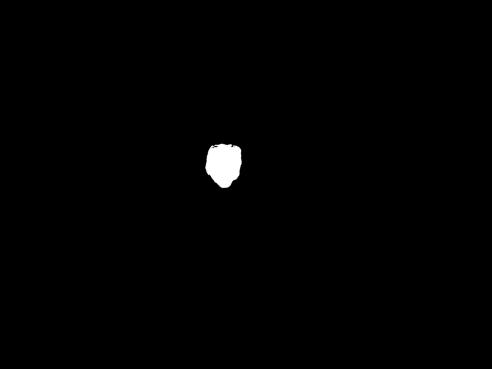} \\

\end{tabular}
\\[1ex]
\caption{Qualitative results, showing for each image the Ground Truth mask of the manipulated region and the corresponding predictions of the MMFusion approach as well as of the CMX architecture that uses each filter separately. Source dataset of each image: rows 1-2: Coverage; rows 3-4: Columbia; rows 5-6: Casiav1+; rows 7-8: CocoGlide; rows 9-10: DSO-1.}
\label{table:qual_ablation}
\end{figure*}

In this section, for the purpose of contrasting the employed forensic filters (SRM, Bayar conv, NoisePrint++), we train a dual-branch CMX architecture where each filter's output serves as the single auxiliary input alongside the RGB image. The outcomes are presented in Table \ref{table:ablation}, along with the number of parameters (in millions) and runtime (for a single image on an RTX 3090 GPU) for all methods. During this training the Bayar convolutional layer is trainable, while SRM and NoisePrint++ are kept frozen. We can see that NoisePrint++'s editing-history-based training helps achieve the best performance on DSO-1, where manipulations are covered using post-processing operations, while SRM and Bayar perform better in CocoGlide and Coverage. Coverage contains only copy-move manipulations for which NoisePrint's++ camera model identification might not provide robust enough forensic traces, whereas CocoGlide's manipulations are diffusion-based inpaintings potentially resulting in distinct artifacts that diverge from conventional editing histories. Consequently, NoisePrint++ encounters difficulties in effectively handling such cases. We also compare all methods that use a single forensic filter to our multi-modal fusion approaches and we can see that the latter effectively combine the forensic traces provided by the different filters, resulting in increased performance. To substantiate our rationale for introducing shared weights between RGB branches in order to enhance regularization within the late fusion paradigm, we additionally evaluate a variation of our method that does not employ weight sharing, and we observe that weight-sharing does contribute to improved performance.

We also qualitatively compare our proposed MMFusion approach with the dual-branch CMX architecture that uses each filter separately in Fig \ref{table:qual_ablation}, and we confirm the ability of MMFusion to effectively combine the information of each filter. More specifically, for the examples from the Coverage dataset (rows 1 and 2), MMFusion accurately localizes the manipulated region that is shown in the ground truth mask, while the SRM and Noiseprint++ models either over-estimate or miss crucial parts. In the splicing samples from the Columbia dataset (rows 3 and 4), the MMFusion-generated localization predictions are the least noisy ones among the results of all compared approaches. On the images of the Casiav1+ dataset (rows 5 and 6), MMFusion correctly identifies the manipulated region of the woman in the red dress, including most of the parts of her arms that other methods missed, and avoiding the incorrect detection of the person playing golf by the Bayar and SRM filters. Similar comments can be made by looking at the CocoGlide images (rows 7 and 8), where, for the bus example Noiseprint++ and Bayar leave gaps in the prediction while in the other image they outright miss detecting some of the donuts as manipulated, contrary to MMFusion that gives the closest localization result to the ground truth. Closing with some samples from the DSO-1 dataset (rows 9 and 10), we observe that Bayar and SRM over-estimate manipulations in the presence of people and faces, with our approach gives the cleanest and most accurate result.

\subsection{Evaluation and Comparisons on Video Manipulation Datasets}

We further evaluate our proposed method MMFusion on the VCMS, VPVM, VPIM \cite{videofact} video datasets, which contain manipulated videos, and compare our approach with the state-of-the-art video manipulation detection model VideoFACT \cite{videofact} and, by extension, with image manipulation detection methods that were used for comparisons in \cite{videofact}. As reported in the latter, VideoFACT was trained on a combination of the VCMS, VPVM and VPIM video datasets and three image datasets. For MMFusion, we assess models of it trained on different datasets: a model that is trained on image data only (i.e. the same trained model that is evaluated in the preceding sections), a model that is trained only on the training splits of the employed video datasets (VCMS, VPVM, VPIM), and a model that is trained on all of our training data (i.e. all images and video frames of Table \ref{table:data}). The results are presented in Tables \ref{table:video_loc_results} and \ref{table:video_det_results}.

We observe that even without training on the video datasets, our model reaches state-of-the-art performance on the VCMS dataset that contains splicing manipulations while it outperforms other IMLD methods on the VPVM and VPIM datasets, thus establishing a new baseline of VMLD performance for IMLD models. Our models trained on video-only and both image-video data also achieve state-of-the-art performance across all datasets, showcasing that designing complex temporal modules that are also computationally expensive is not necessary to achieve state-of-the-art performance on VMLD tasks; simply training on video data should suffice, at least for the types of video manipulations present on our evaluation datasets. This also exposes the need for new more sophisticated VMLD datasets that contain manipulations with more complex temporal elements, in order to be able to more accurately compare VMLD models.

\begin{table}[ht]
\centering
 \caption{Average pixel-level F1 scores for the video localization task on the considered models and datasets. Best (higher) scores in bold and second best scores underlined. Results for all models except for the proposed ones are taken from \cite{videofact}.}
 \label{table:video_loc_results}
\begin{tabular}{ | c|| c  c  c | c | }
 \hline
 Model & VCMS & VPVM & VPIM & AVG \\
 \hline \hline

VIDEOFACT\cite{videofact}  & 0.526      & 0.697      & 0.547  & 0.590\\
NoisePrint\cite{noiseprint}  & 0.030      & 0.013      & 0.010  & 0.018\\
MantraNet\cite{mantranet}  & 0.114      & 0.145      & 0.064  & 0.108\\
MVSS-Net\cite{mvss}  & 0.557      & 0.279      & 0.042  & 0.293\\
\hline

MMFusion (Image Data) & 0.838 & 0.520 & 0.229  &  0.529 \\
MMFusion (Video Data) & \textbf{0.952} & \textbf{0.945} & \textbf{0.686}  &  \textbf{0.861} \\
MMFusion (All data) & \underline{0.921} &  \underline{0.898} & \underline{0.579}  &  \underline{0.799} \\
\hline
\end{tabular}
\end{table}

\begin{table}[ht]
\centering
 \caption{Balanced accuracy scores for the video detection task on the considered models and datasets. Best (higher) scores in bold and second best scores underlined. Results for all models except for the proposed ones are taken from \cite{videofact}.}
 \label{table:video_det_results}
\begin{tabular}{ | c|| c  c  c | c | }
 \hline
 Model & VCMS & VPVM & VPIM & AVG \\
 \hline \hline

VIDEOFACT\cite{videofact}  & \textbf{0.987}      & \underline{0.950}      & 0.797  & \underline{0.911}\\
NoisePrint\cite{noiseprint}  & 0.500      & 0.500      & 0.500  & 0.500\\
MantraNet\cite{mantranet}  & 0.500      & 0.500      & 0.500  & 0.500\\
MVSS-Net\cite{mvss}  & 0.602      & 0.529      & 0.492  & 0.541\\
\hline

MMFusion (Image Data) & 0.915 & 0.650 & 0.510  &  0.692 \\
MMFusion (Video Data)  & \underline{0.963} & \textbf{0.962} & \textbf{0.878}  &  \textbf{0.934} \\
MMFusion (All data) & 0.923 & 0.923 & \underline{0.822} & 0.889\\
\hline
\end{tabular}
\end{table}

\subsection{Robustness Analysis}

\begin{figure*}[h!]
\centering
\includegraphics[width=0.8\textwidth]{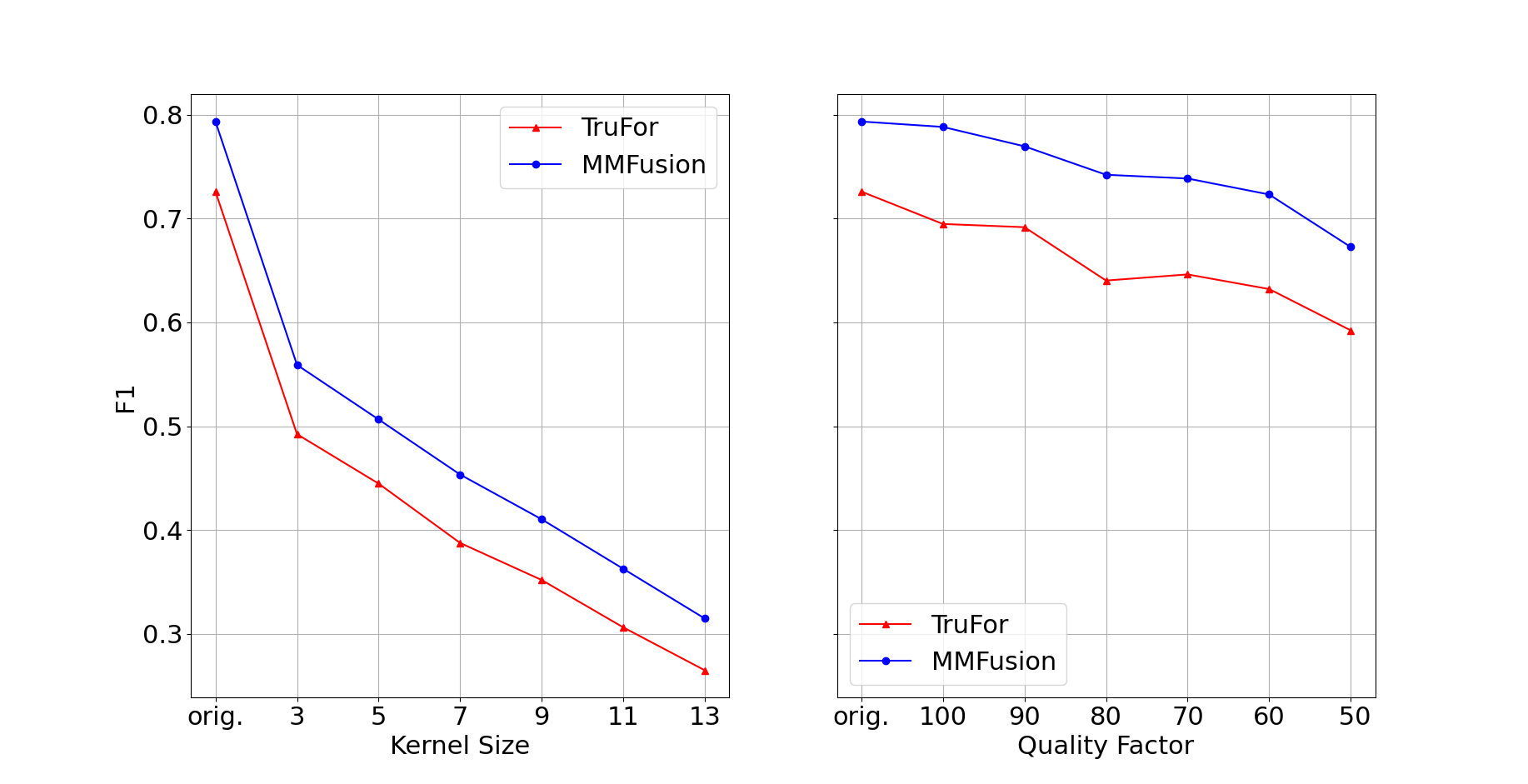}
\caption{Robustness analysis with regards to Gaussian blur (left) and JPEG compression (right). Higher F1 values are better.} \label{ra}
\end{figure*}

In this section, we include experiments performed on images with varying quality degradations to demonstrate the robustness of our approach, similarly to Guillaro et al \cite{trufor}. We use the Casiav1+ dataset and perform Gaussian blurring with different kernel sizes (3, 5, 7, 9, 11, 13) and JPEG compression with varying quality factors (100, 90, 80, 70, 60, 50) and compare the emerging pixel-level F1 scores to our baseline, TruFor. The findings depicted in Fig. \ref{ra} demonstrate that our MMfusion architecture exhibits good robustness across a broad spectrum of degradations, maintaining a consistent advantage over TruFor.

\section{Explainability of IMLD models: Quantifying the importance of different forensic filters}

\newcommand{\colorcell}[1]{
  \ifdim #1pt < 0pt
    \cellcolor{red!50}#1
  \else
    \ifdim #1pt < 0.06pt
      \cellcolor{yellow!50}#1
    \else
      \cellcolor{green!50}#1
    \fi
  \fi
}

\newcommand{\colorcellbold}[1]{
  \ifdim #1pt < 0pt
    \cellcolor{red!50}\textbf{#1}
  \else
    \ifdim #1pt < 0.06pt
      \cellcolor{yellow!50}\textbf{#1}
    \else
      \cellcolor{green!50}\textbf{#1}
    \fi
  \fi
}

\newcommand{\colorcellunderline}[1]{
  \ifdim #1pt < 0pt
    \cellcolor{red!50}\underline{#1}
  \else
    \ifdim #1pt < 0.06pt
      \cellcolor{yellow!50}\underline{#1}
    \else
      \cellcolor{green!50}\underline{#1}
    \fi
  \fi
}

\begin{table*}[h!]
\centering
 \caption{Average drop in the pixel-level F1 scores (calculated using the ground truth mask for the localization task), before and after masking each modality with either zeros (above the diving horizontal line) or with another random image (below the diving horizontal line). Best (higher) scores in bold and second best scores underlined.}
 \label{table:explanations}
\begin{tabular}{ | c|| c  c  c  c  c | c | }
 \hline
 Masked Modality $\rightarrow$ Mask Type & Coverage & Columbia & Casiav1+ & CocoGlide & DSO-1 & AVG \\
 \hline \hline

NoisePrint $\rightarrow$ 0& \colorcellunderline{0.0225}      & \colorcellunderline{0.0236}      & \colorcellunderline{0.0168}      & \colorcellunderline{-0.0222}      & \colorcellbold{0.2015}  & \colorcellunderline{0.0484}\\
Bayar Conv $\rightarrow$ 0&\colorcell{-0.0026}    & \colorcell{-0.0051}      & \colorcell{0.0154}      & \colorcell{-0.0442}      & \colorcell{0.0253}  & \colorcell{-0.0022}\\
SRM        $\rightarrow$ 0& \colorcellbold{0.2367}
    & \colorcellbold{0.0276}      & \colorcellbold{0.0649}     & \colorcellbold{0.1376}      & \colorcellunderline{0.0709}  & \colorcellbold{0.1075}\\
\hline
NoisePrint $\rightarrow$ Random Image& \colorcell{0.0225}& \colorcellunderline{0.0155}  & \colorcellunderline{0.0200} & \colorcell{-0.0268}    & \colorcellbold{0.2277}  & \colorcellunderline{0.0487}\\
Bayar Conv $\rightarrow$ Random Image& \colorcellunderline{0.0267}& \colorcell{-0.0041}  & \colorcell{0.0175}  & \colorcellunderline{-0.0072}  & \colorcell{0.0385}  & \colorcell{0.0143}\\
SRM        $\rightarrow$ Random Image& \colorcellbold{0.2269} & \colorcellbold{0.0393}   & \colorcellbold{0.0677}   & \colorcellbold{0.1505}  & \colorcellunderline{0.1487}  & \colorcellbold{0.1266}\\
\hline
\end{tabular}
\end{table*}

\begin{table*}[h!]
\centering
 \caption{Drop in Prediction Quality (PQ) measure scores using the output of the unmasked prediction as ground truth mask for the localization task, after masking each modality with zeros (above the diving horizontal line) or with another random image (below the diving horizontal line). Best (lower) scores in bold and second best scores underlined.}
 \label{table:explanations_blind}
\begin{tabular}{ | c|| c  c  c  c  c | c | }
 \hline
 Masked Modality $\rightarrow$ Mask Type & Coverage & Columbia & Casiav1+ & CocoGlide & DSO-1 & AVG \\
 \hline \hline

NoisePrint $\rightarrow$ 0& 0.8215 & \underline{0.9381}      & 0.8605      & 0.5689      & \textbf{0.7349}  &  \underline{0.7848} \\
Bayar Conv $\rightarrow$ 0& \underline{0.8079} & 0.9747      & \underline{0.8291}      & \underline{0.5412}      & 0.9576  & 0.8221 \\
SRM        $\rightarrow$ 0& \textbf{0.5521}& \textbf{0.9356}      & \textbf{0.7934}      & \textbf{0.3636}      & \underline{0.9078}  & \textbf{0.7105}\\
\hline
NoisePrint $\rightarrow$ Random Image& 0.8340 & \underline{0.9489}  & 0.8491 & 0.5889  & \textbf{0.7117}  & \underline{0.7865} \\
Bayar Conv $\rightarrow$ Random Image& \underline{0.7892}& 0.9726  & \underline{0.8299} & \underline{0.5272}  & 0.9317  & 0.8101 \\
SRM        $\rightarrow$ Random Image& \textbf{0.5455}& \textbf{0.9231}  & \textbf{0.7960} & \textbf{0.3370}  & \underline{0.8051}  & \textbf{0.6813}\\
\hline
\end{tabular}

\end{table*}

Image manipulation localization and detection as a Machine Learning task is inherently explainable to some extent, as the localization map predicted can serve as the explanation for the detection prediction (e.g. a ``manipulated'' classification decision can be explained by ``the region shown in this localization map is predicted to be manipulated'', in an analogy to the form of explanations produced by e.g. T-TAME \cite{ttame} for ImageNet classifiers). Despite this, we investigate how to extend the explainability of our multi-modal models for IMLD, given the high importance for end-users to be offered rich insights about the decision-making processes, to allow for greater transparency and trustworthiness of the classification decisions. This investigation builds on our preceding ablation study, which revealed that different forensic filters exhibit complementary performance characteristics (Sec. \ref{ablation}). The importance of each filter becomes apparent when evaluating its effectiveness against distinct types of manipulated images; for instance, some filters are particularly adept at identifying copy-move forgeries seen in the Coverage dataset, while others excel with splicing forgeries from the Columbia dataset, which lacks post-processing. To dive deeper into this, we employ a perturbation-based explanation method, in the spirit of methods like LIME and SHAP \cite{lime,shap}: we mask one modality (i.e. filter output) at each time and observe MMFusion's resulting drop in performance. Essentially, we replace the input of the chosen filter with either zeros or a random pristine image and quantify the importance of the filter as the drop in localization F1 (higher drop means the filter is more important). This approach highlights the filter's reliance on specific data features. The results are displayed in Table \ref{table:explanations}. Our tests across different datasets, each exhibiting different manipulations (as detailed in Sec. \ref{sec:testdatasets}), demonstrate that each filter is tailored to detect particular forgery characteristics, thereby offering a comprehensive analysis framework for diverse forensic scenarios.

\begin{figure*}
\centering
\begin{subfigure}{.3\textwidth}
  \centering
  \includegraphics[width=.8\linewidth]{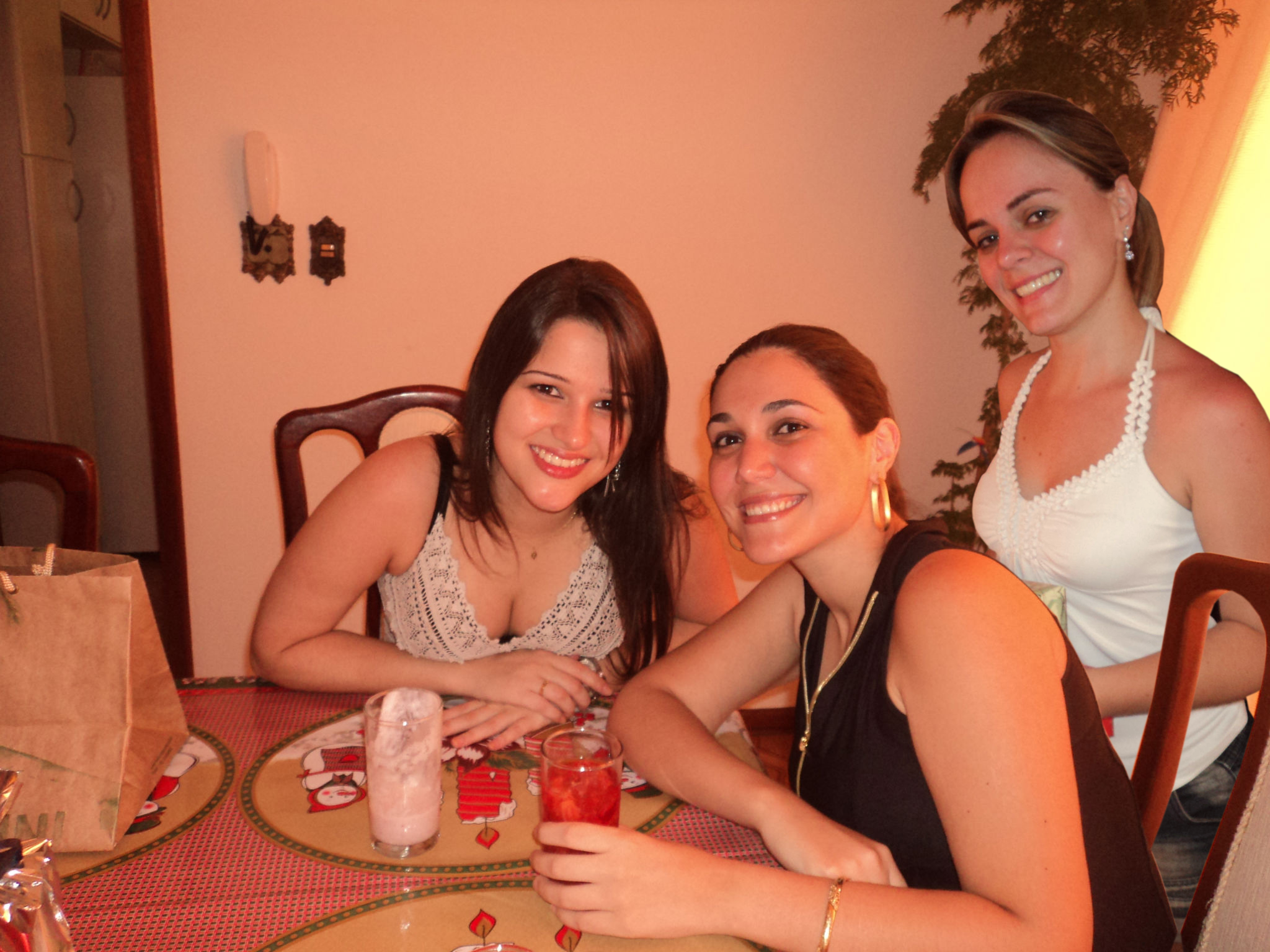}
  \caption{Image w/ splicing forgery}
  \vspace{8pt}
\end{subfigure}
\begin{subfigure}{.3\textwidth}
  \centering
  \includegraphics[width=.8\linewidth]{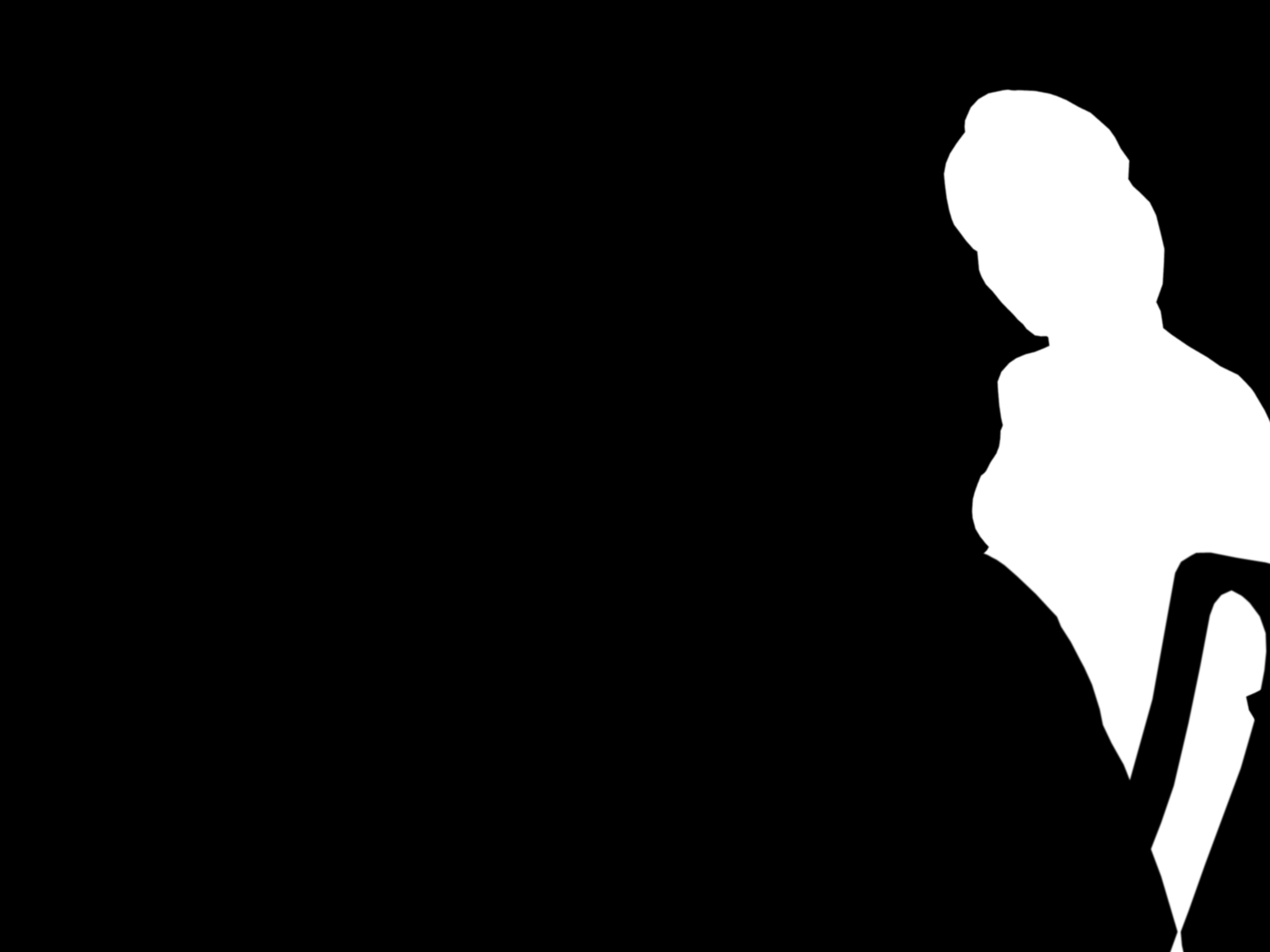}
  \caption{Ground Truth}
  \vspace{8pt}
\end{subfigure}
\begin{subfigure}{.3\textwidth}
  \centering
  \includegraphics[width=.8\linewidth]{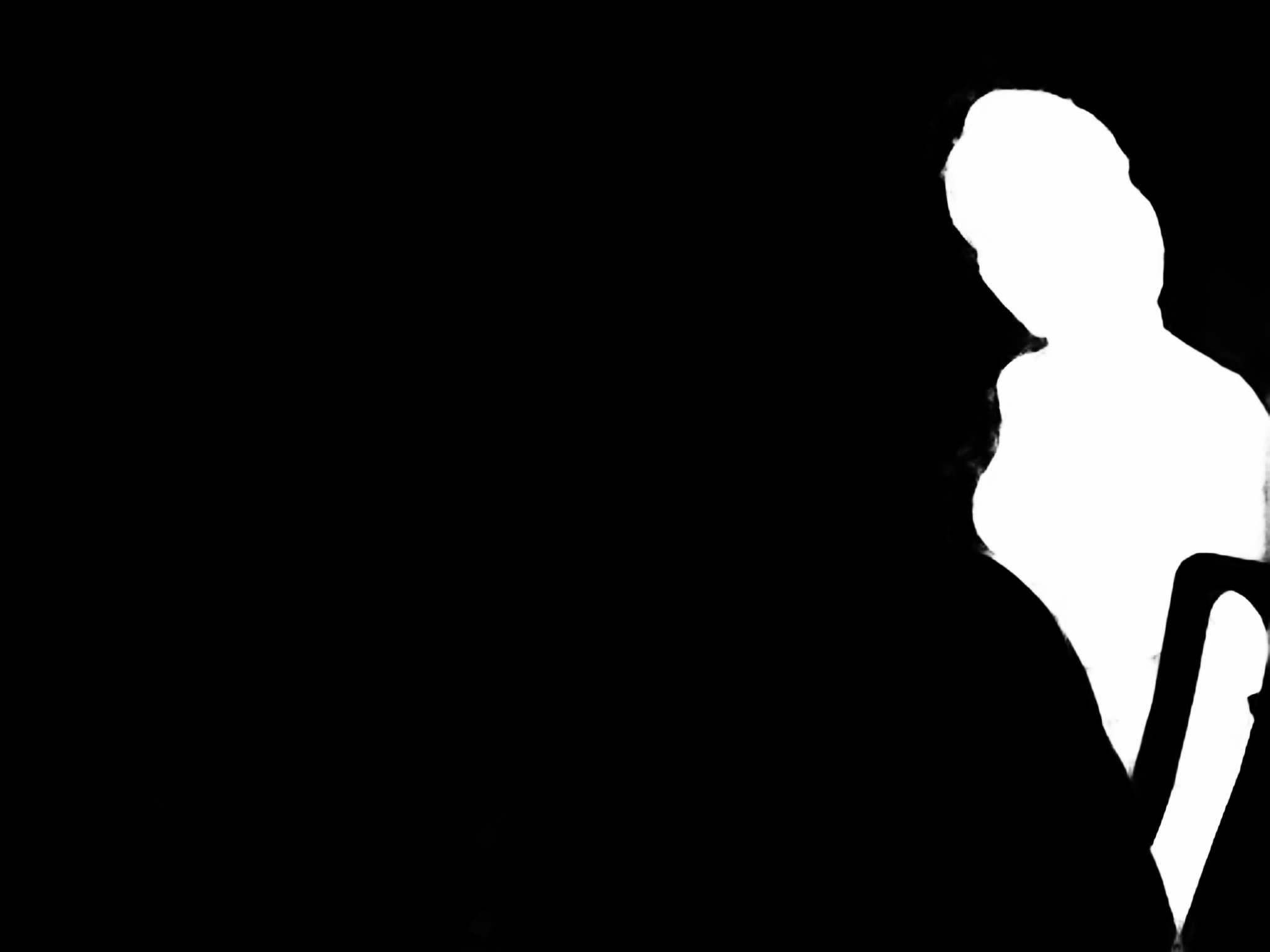}
  \caption{MMFusion prediction}
  \vspace{8pt}
\end{subfigure}
\begin{subfigure}{.3\textwidth}
  \centering
  \includegraphics[width=.8\linewidth]{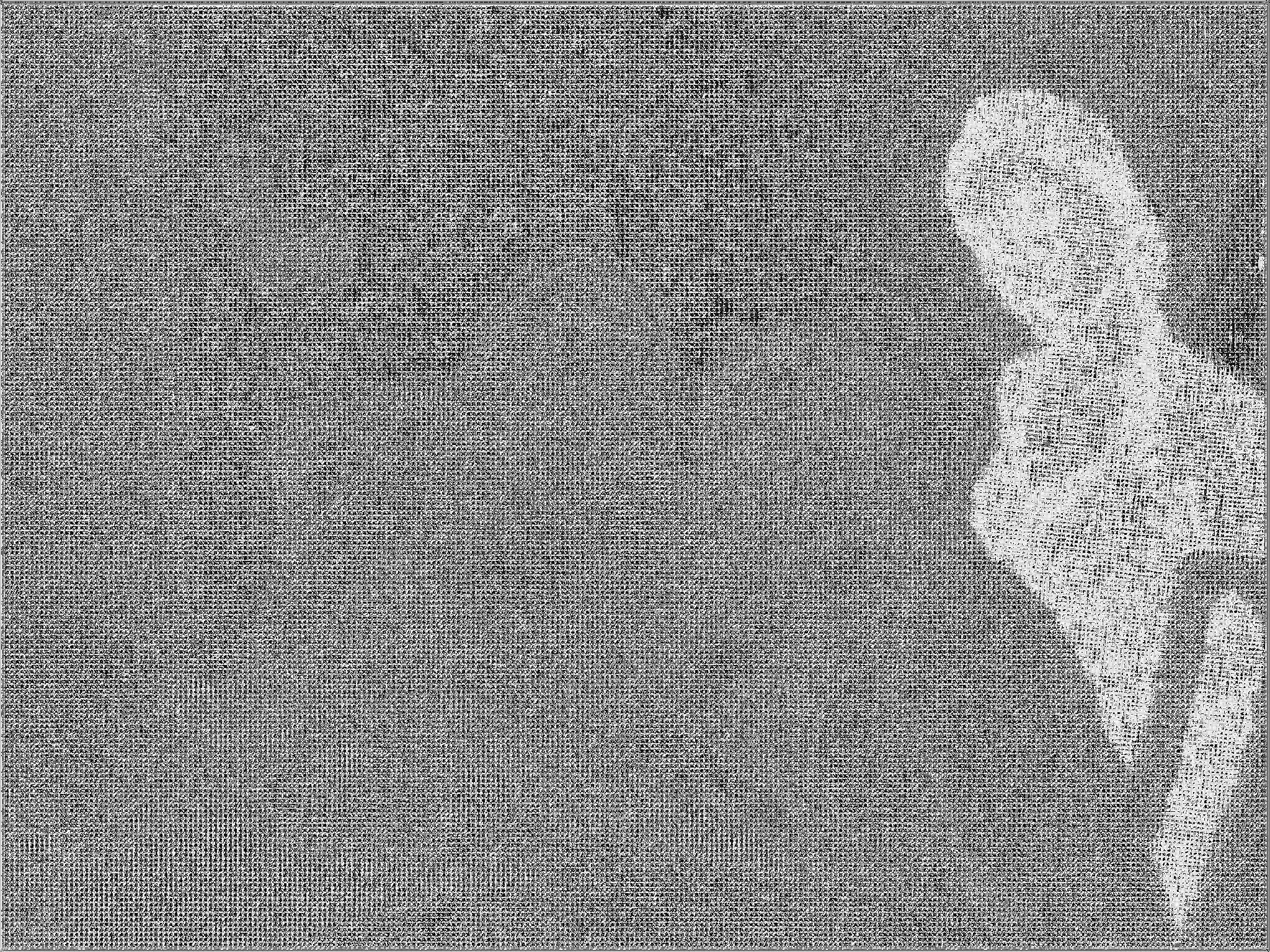}
  \caption{NoisePrint++ output}
\end{subfigure}
\begin{subfigure}{.3\textwidth}
  \centering
  \includegraphics[width=.8\linewidth]{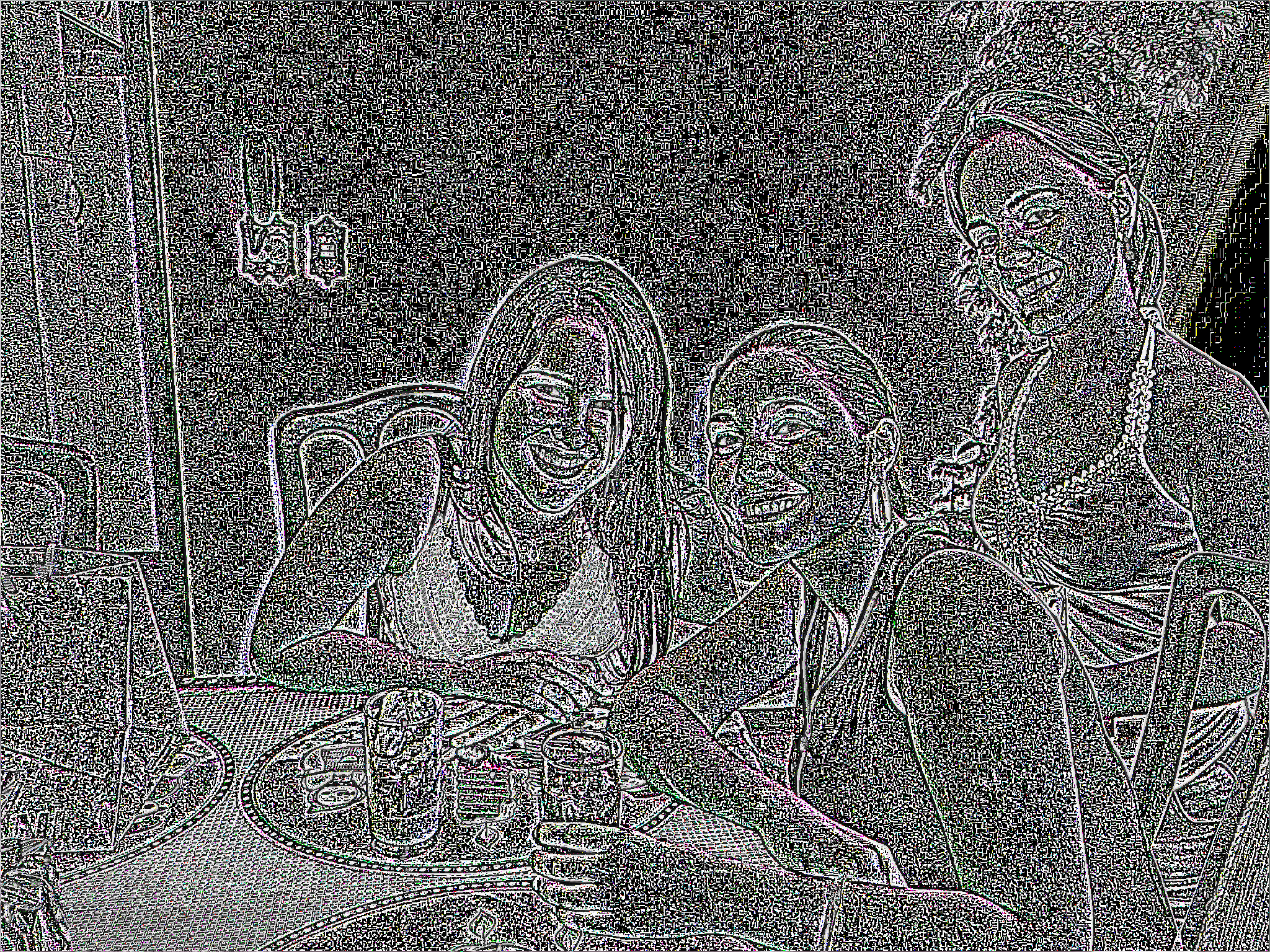}
  \caption{Bayar Conv output}
\end{subfigure}
\begin{subfigure}{.3\textwidth}
  \centering
  \includegraphics[width=.8\linewidth]{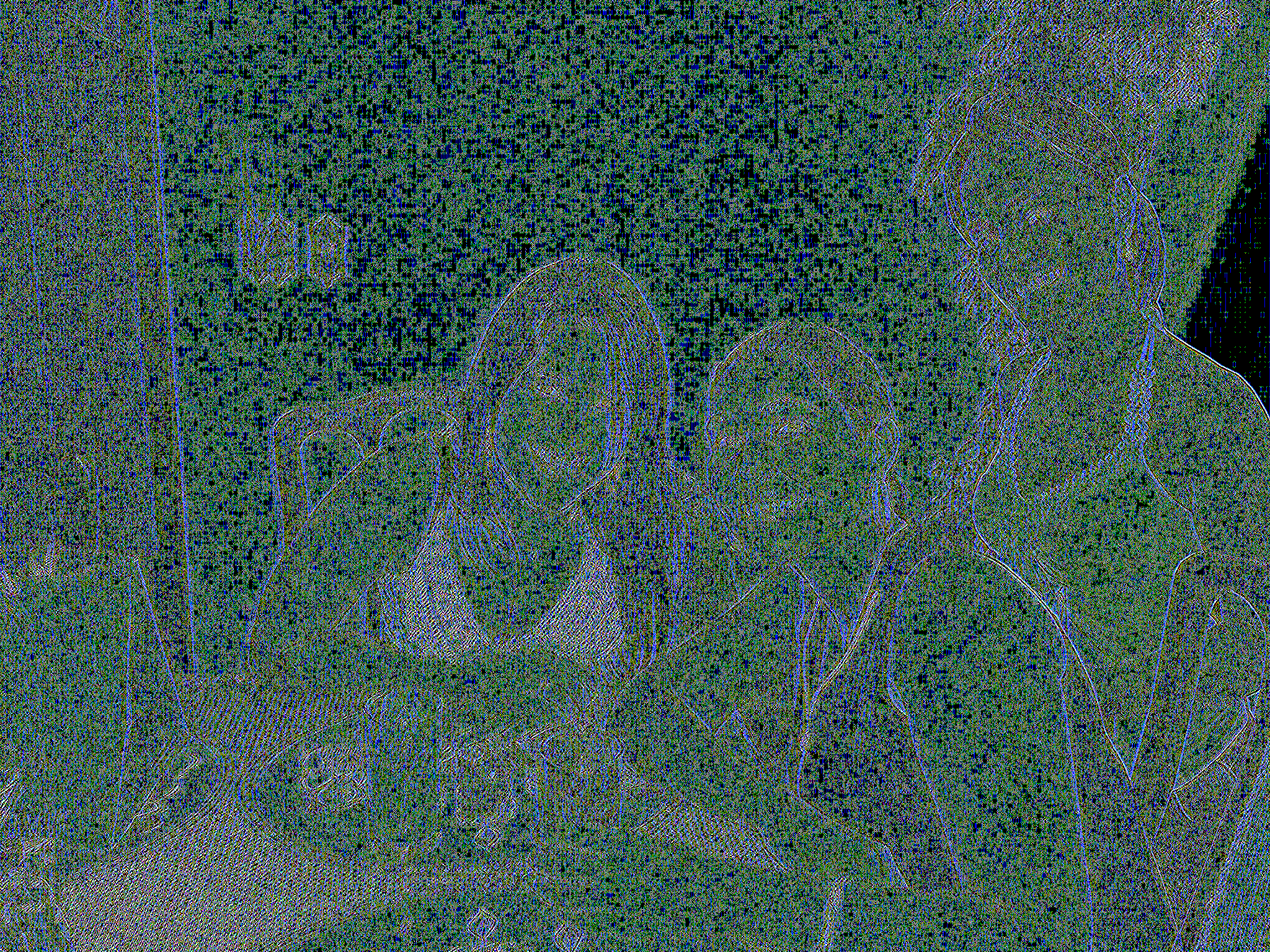}
  \caption{SRM output}
\end{subfigure}
\vspace{5pt}
\caption{Visualization of the filter outputs for a DSO-1 dataset image. Top row: (a) Input image with splicing forgery, (b) Ground Truth mask of the manipulated region, (c) Prediction/Detection mask of MMFusion. Bottom row: (d)-(f) Output of each filter.}
\label{fig:np_imp}
\end{figure*}

\begin{figure*}
\centering
\begin{subfigure}{.3\textwidth}
  \centering
  \includegraphics[width=.8\linewidth]{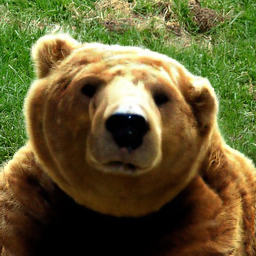}
  \caption{Image w/ inpainting forgery}
  \vspace{8pt}
\end{subfigure}
\begin{subfigure}{.3\textwidth}
  \centering
  \includegraphics[width=.8\linewidth]{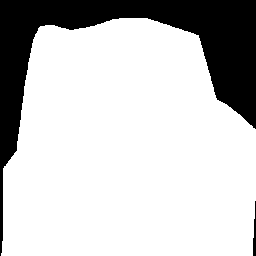}
  \caption{Ground Truth}
  \vspace{8pt}
\end{subfigure}
\begin{subfigure}{.3\textwidth}
  \centering
  \includegraphics[width=.8\linewidth]{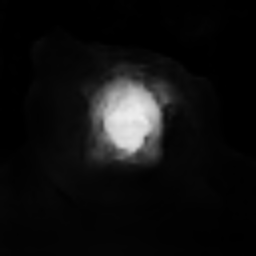}
  \caption{MMFusion prediction}
  \vspace{8pt}
\end{subfigure}
\begin{subfigure}{.3\textwidth}
  \centering
  \includegraphics[width=.8\linewidth]{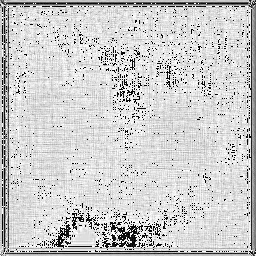}
  \caption{NoisePrint++ output}
\end{subfigure}
\begin{subfigure}{.3\textwidth}
  \centering
  \includegraphics[width=.8\linewidth]{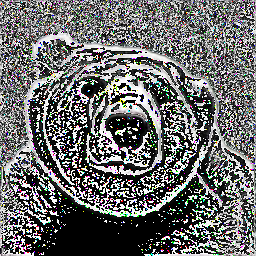}
  \caption{Bayar Conv output}
\end{subfigure}
\begin{subfigure}{.3\textwidth}
  \centering
  \includegraphics[width=.8\linewidth]{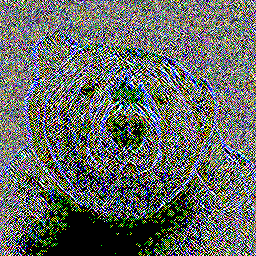}
  \caption{SRM output}
\end{subfigure}
\vspace{5pt}
\caption{Visualization of the filter outputs for a CocoGlide dataset image. Top row: (a) Input image with inpainting forgery, (b) Ground Truth mask of the manipulated region, (c) Prediction/Detection mask of MMFusion. Bottom row: (d)-(f) Output of each filter.}
\label{fig:bayar_imp}
\end{figure*}

\begin{figure*}
\centering
\begin{subfigure}{.3\textwidth}
  \centering
  \includegraphics[width=.8\linewidth]{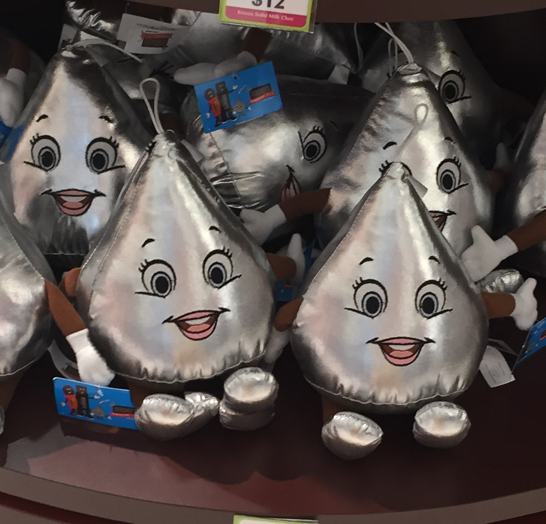}
  \caption{Image w/ copy-move forgery}
  \vspace{8pt}
\end{subfigure}
\begin{subfigure}{.3\textwidth}
  \centering
  \includegraphics[width=.8\linewidth]{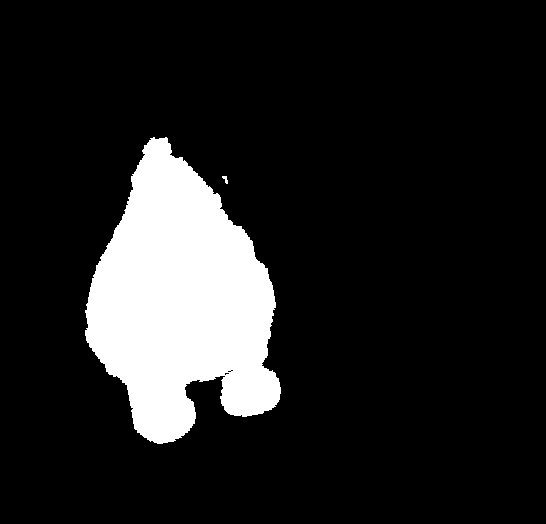}
  \caption{Ground Truth}
  \vspace{8pt}
\end{subfigure}
\begin{subfigure}{.3\textwidth}
  \centering
  \includegraphics[width=.8\linewidth]{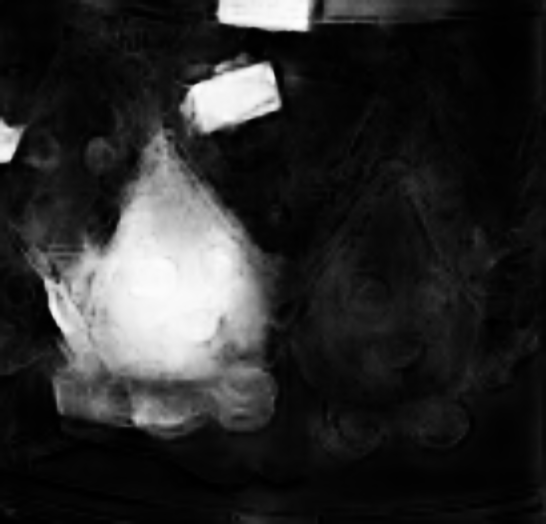}
  \caption{MMFusion prediction}
  \vspace{8pt}
\end{subfigure}
\begin{subfigure}{.3\textwidth}
  \centering
  \includegraphics[width=.8\linewidth]{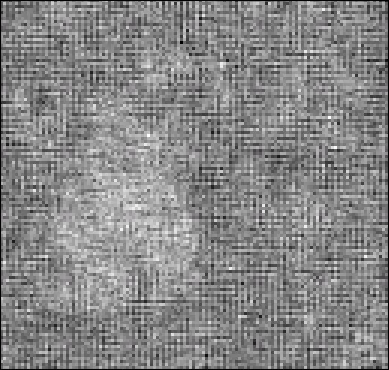}
  \caption{NoisePrint++ output}
\end{subfigure}
\begin{subfigure}{.3\textwidth}
  \centering
  \includegraphics[width=.8\linewidth]{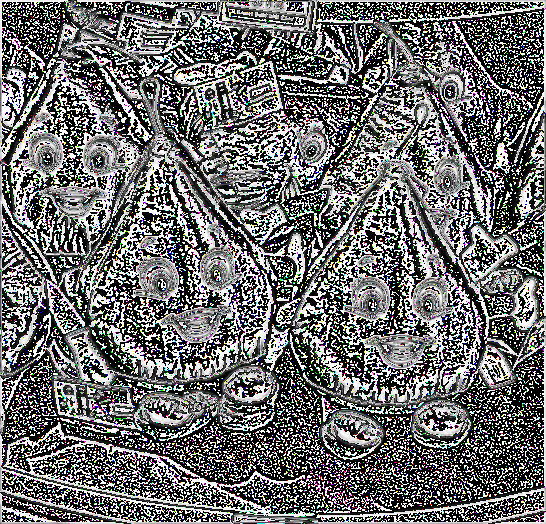}
  \caption{Bayar Conv output}
\end{subfigure}
\begin{subfigure}{.3\textwidth}
  \centering
  \includegraphics[width=.8\linewidth]{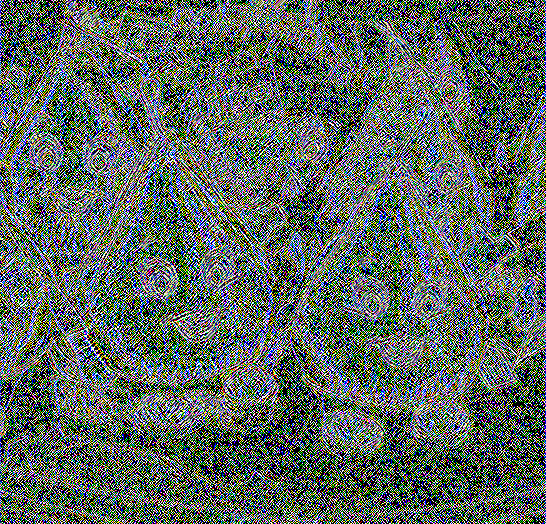}
  \caption{SRM output}
\end{subfigure}
\vspace{5pt}
\caption{Visualization of the filter outputs for a Coverage dataset image. Top row: (a) Input image with inpainting forgery, (b) Ground Truth mask of the manipulated region, (c) Prediction/Detection mask of MMFusion. Bottom row: (d)-(f) Output of each filter.} 
\label{fig:srm_imp}
\end{figure*}

This explanation method, however, relies on our knowledge of the ground truth mask for the image. In order to be able to provide explanations for in-the-wild images, we expand our methodology and propose Drop in Prediction Quality (PQ) as a new evaluation measure. Drop in Prediction Quality is calculated as the F1 measure for the masked prediction using the original unmasked prediction as the ground truth. For the new PQ measure, lower value means the masked modality is more important. We present the blind explanations results in Table \ref{table:explanations_blind}. We observe that these results are consistent with those reported in Table \ref{table:explanations} in highlighting SRM and NoisePrint++ as the most important overall filters. Specifically, the SRM filter is the most important for copy-move and inpainting forgeries, whereas NoisePrint++ is very helpful for recognizing splicing forgeries. Our hypothesis posits that SRM's effectiveness stems from its sensitivity to edge artifacts produced during the manipulation process, while NoisePrint++ excels at identifying differences in texture between the source and target image that exist in splicing forgeries. To provide more insight on what the output of each filter looks like, a few examples of filter outputs for images with different manipulations are illustrated in Figures \ref{fig:np_imp}, \ref{fig:bayar_imp} and \ref{fig:srm_imp}.

\section{Conclusion}

In this work, we expand an existing encoder-decoder architecture for image manipulation localization and detection (IMLD) to support multiple forensic filters as inputs. We examine two filter fusion paradigms: one that generates independent features from each forensic filter before fusing them (late fusion), and another that performs early mixing of modal outputs to produce combined features (early fusion). By leveraging three forensic filters, i.e. Bayar convolution, SRM and NoisePrint++, we show that these filters provide distinct and complementary forensic capabilities and can be effectively combined, as hypothesized. We then introduce a feature re-weighting decoder and deploy it alongside early fusion to propose the MMFusion architecture. Extensive experiments demonstrate that MMFusion achieves state-of-the-art performance across multiple image datasets, showcasing good generalization and robustness, and its effectiveness in leveraging diverse forensic artifacts from different filters. Additionally, we apply MMFusion to video manipulation datasets, also reaching state-of-the-art performance. Finally, we further assess the contribution of each forensic filter to the MMFusion model's decisions.

\newpage

\vfill

\end{document}